\documentclass[10pt,twocolumn,letterpaper]{article}

\usepackage{cvpr}              

\usepackage{graphicx}
\usepackage{amsmath}
\usepackage{amssymb}
\usepackage{booktabs}

\usepackage[pagebackref,breaklinks,colorlinks]{hyperref}

\usepackage[capitalize]{cleveref}
\crefname{section}{Sec.}{Secs.}
\Crefname{section}{Section}{Sections}
\Crefname{table}{Table}{Tables}
\crefname{table}{Tab.}{Tabs.}

\def\cvprPaperID{8282} 
\def\confName{CVPR}
\def\confYear{2023}

\usepackage{amsmath,amsfonts,bm}

\def\eqref#1{equation~\ref{#1}}

\def\1{\bm{1}}

\def\vs{{\bm{s}}}

\def\vv{{\bm{v}}}

\DeclareMathAlphabet{\mathsfit}{\encodingdefault}{\sfdefault}{m}{sl}
\SetMathAlphabet{\mathsfit}{bold}{\encodingdefault}{\sfdefault}{bx}{n}

\newcommand{\R}{\mathbb{R}}

\usepackage{hyperref}
\usepackage{url}

\usepackage[utf8]{inputenc} 
\usepackage[T1]{fontenc}    
\usepackage{hyperref}       
\usepackage{url}            
\usepackage{amsfonts}       
\usepackage{nicefrac}       
\usepackage{microtype}      
\usepackage{xcolor}         

\usepackage{float}
\usepackage{multirow}
\usepackage{amsfonts}
\usepackage{dsfont}
\usepackage{enumitem}
\usepackage{xcolor}
\usepackage{array}
\usepackage{makecell}
\usepackage[export]{adjustbox}
\usepackage{hyperref}
\usepackage{ifthen}
\usepackage{wrapfig}
\usepackage{arydshln}
\newcommand{\encoder}[1]{$E$}
\newcommand{\decoder}[1]{$D$}
\newcommand{\maskoutputv}[1]{$\hat{x}_v$}
\newcommand{\methodname}[1]{UDOP}
\newcommand{\rvlcdip}[1]{RVL-CDIP}
\newcommand{\publaynet}[1]{PubLayNet}
\newcommand{\tablebank}[1]{TableBank}
\newcommand{\docbank}[1]{DocBank}
\newcommand{\websrc}[1]{WebSRC}
\newcommand{\visualmrc}[1]{VisualMRC}
\newcommand{\iitcdip}[1]{IIT-CDIP}
\newcommand{\iitcdipfull}[1]{IIT-CDIP Test Collection 1.0}
\newcommand{\funsd}[1]{FUNSD}
\newcommand{\cord}[1]{CORD}
\newcommand{\duebenchmark}[1]{DUE-Benchmark}

\newcommand{\docvqa}[1]{DocVQA}
\newcommand{\infovqa}[1]{InfographicsVQA}
\newcommand{\kleister}[1]{Kleister Charity}
\newcommand{\pwc}[1]{PWC}
\newcommand{\deepform}[1]{DeepForm}
\newcommand{\wtq}[1]{WTQ}
\newcommand{\tabfact}[1]{TabFact}

\newcommand{\dual}[1]{\methodname{}-Dual}
\newcolumntype{P}[1]{>{\centering\arraybackslash}p{#1}}

\title{
Unifying Vision, Text, and Layout for Universal Document Processing
}

\author{
Zineng Tang$^{1,2}$\qquad
Ziyi Yang$^{2}\thanks{Corresp. authors: ziyiyang@microsoft.com, mbansal@cs.unc.edu}$\qquad
Guoxin Wang$^3$\qquad
Yuwei Fang$^2$\qquad
Yang Liu$^2$\qquad\\
Chenguang Zhu$^2$\qquad
Michael Zeng$^2$\qquad
Cha Zhang$^3$\qquad
Mohit Bansal$^{1*}$
\\
$^1$University of North Carolina at Chapel Hill\\
$^2$Microsoft Azure Cognitive Services Research\\
$^3$Microsoft Azure Visual Document Intelligence
}

\begin{document}

\maketitle

\begin{abstract}
We propose Universal Document Processing (\methodname{}), a foundation Document AI model which unifies text, image, and layout modalities together with varied task formats, including document understanding and generation. \methodname{} leverages the spatial correlation between textual content and document image to model image, text, and layout modalities with one uniform representation. With a novel Vision-Text-Layout Transformer, \methodname{} unifies pretraining and multi-domain downstream tasks into a prompt-based sequence generation scheme. \methodname{} is pretrained on both large-scale unlabeled document corpora using innovative self-supervised objectives and diverse labeled data. \methodname{} also learns to generate document images from text and layout modalities via masked image reconstruction. To the best of our knowledge, this is the first time in the field of document AI that one model simultaneously achieves high-quality neural document editing and content customization.
Our method sets the state-of-the-art on 8 Document AI tasks, e.g., document understanding and QA, across diverse data domains like finance reports, academic papers, and websites. \methodname{} ranks first on the leaderboard of the Document Understanding Benchmark.\footnote{Code and models: \url{https://github.com/microsoft/i-Code/tree/main/i-Code-Doc}}
\end{abstract}

\begin{figure*}[t]
  \centering
  \includegraphics[width=0.99\textwidth]{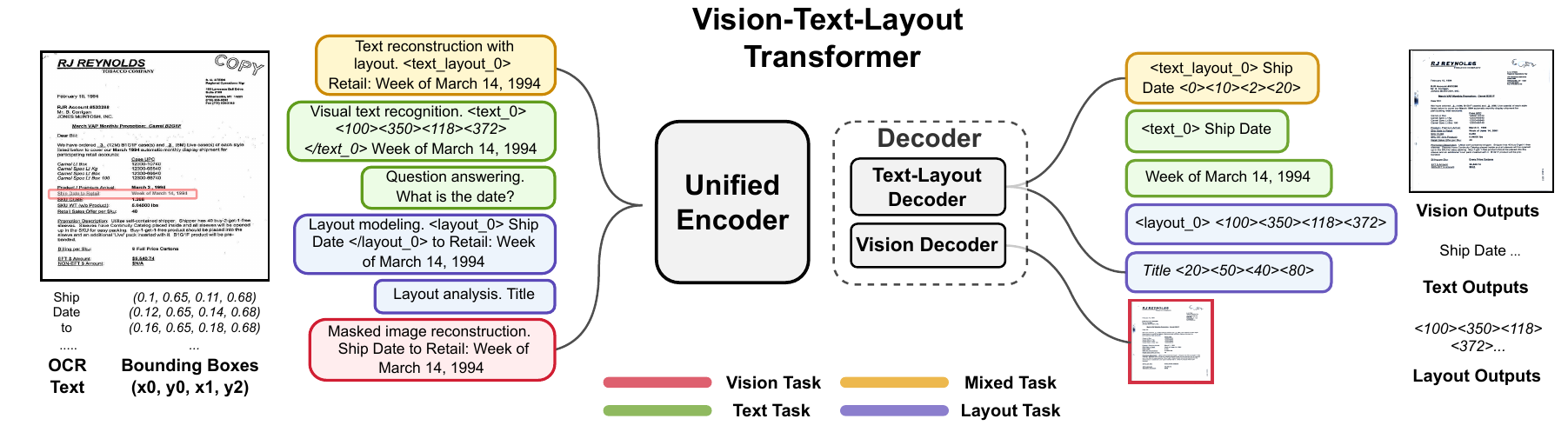}
  \caption{
    \methodname{} unifies vision, text, and layout through vision-text-layout Transformer and unified generative pretraining tasks including vision task, text task, layout task, and mixed task. We show the task prompts (left) and task targets (right) for all self-supervised objectives (joint text-layout reconstruction, visual text recognition, layout modeling, and masked autoencoding) and two example supervised objectives (question answering and layout analysis).
}
\label{fig:architecture}
\end{figure*}

\section{Introduction}
Document Artificial Intelligence studies information extraction, understanding, and analysis of digital documents, e.g., business invoices, tax forms, academic papers, etc. It is a multimodal task where text is structurally embedded in documents, together with other vision information like symbols, figures, and style. Different from classic vision-language research, document data have a 2D spatial layout: text content is structurally spread around in different locations based on diverse document types and formats (e.g., invoices vs. tax forms); formatted data such as figures, tables and plots are laid out across the document.
Hence, effectively and efficiently modeling and understanding the layout is vital for document information extraction and content understanding, for example, title/signature extraction, fraudulent check detection, table processing, document classification, and automatic data entry from documents.

Document AI has unique challenges that set it apart from other vision-language domains. For instance, the cross-modal interactions between text and visual modalities are much stronger here 
than in regular vision-language data, because the text modality is visually-situated in an image. Moreover, downstream tasks are diverse in domains and paradigms, e.g., document question answering \cite{tanaka2021visualmrc}, layout detection\cite{zhong2019publaynet}, classification \cite{harley2015evaluation}, information extraction \cite{li2020docbank}, etc.
This gives rises to two challenges: (1) how to utilize the strong correlation between image, text and layout modalities and unify them to model the document as a whole? (2) how can the model efficiently and effectively learn diverse vision, text, and layout tasks across different domains? 

There has been remarkable progress in Document AI in recent years \cite{pramanik2020towards, xu2020layoutlm, xu2021layoutlmv2, appalaraju2021docformer, garncarek2021lambert, gu2021unidoc, li2021selfdoc, xu2021layoutxlm, li2021structurallm, wu2021lampret, li2021structext, powalski2021going, lee2022formnet, gu2022xylayoutlm, hong2022bros, wang2022lilt, huang2022layoutlmv3}. 
Most of these model paradigms are similar to traditional vision-language frameworks:
one line of work~\cite{pramanik2020towards, xu2020layoutlm, xu2021layoutlmv2, appalaraju2021docformer, gu2021unidoc, li2021selfdoc, xu2021layoutxlm, wu2021lampret, li2021structext, powalski2021going} inherits vision-language models that encode images with a vision network (e.g., vision transformer) and feed the encodings to the multimodal encoder along with text~\cite{tan2019lxmert, li2019visualbert, huang2020pixel, tang2021decembert}; another line of work uses one joint encoder~\cite{Kim2021ViLT, tang2022perceiver} for both text and image~\cite{huang2022layoutlmv3}.
Some models regard documents as text-only inputs~\cite{garncarek2021lambert, li2021structurallm, gu2022xylayoutlm, hong2022bros, wang2022lilt}. In these works, the layout modality is represented as shallow positional embeddings, e.g., adding a 2D positional embedding to text embeddings. The strong correlation between modalities inherent in document data are not fully exploited. Also to perform different tasks, many models have to use task-specific heads, which is inefficient and requires manual design for each task.

To address these challenges, we propose Universal Document Processing (\methodname{}), a foundation Document AI model that unifies vision, text, and layout and different document tasks.
Different from regarding image and document text as two separate inputs in previous works, in \methodname{} we propose to model them with the uniform layout-induced representation (\cref{sec:encoder}): in the input stage, we add embeddings of text tokens  with the features of the image patch where the tokens are located. This simple and novel layout-induced representation greatly enhances the interaction between the text and vision modalities.

Besides the layout-induced representation, to form a uniform paradigm for different vision, text, layout tasks, \methodname{} first builds a homogeneous vocabulary for texts and document layout that converts layout, i.e. bounding boxes, to discretized tokens. Second, we propose Vision-Text-Layout (VTL) Transformer, consisting of a modality-agnostic encoder, text-layout decoder and vision decoder. VTL Transformer allows \methodname{} to jointly encode and decode vision, text, and layout. \methodname{} unites all downstream tasks with a sequence-to-sequence generation framework.

Besides the challenges of modalities unification and task paradigms discussed above, another issue is previous works utilized self-supervised learning objectives that were originally designed for single-modality learning, e.g., masked language modeling, or classical vision-language pretraining, e.g., contrastive learning. We, on the other hand, propose novel self-supervised learning objectives designed to allow holistic document learning, including layout modeling, text and layout reconstruction, and vision recognition that account for text, vision and layout modeling together (\cref{sec:pretrain}). Besides sequential generation, \methodname{} can also generate vision documents by leveraging masked autoencoders (MAE)~\cite{he2021masked} by reconstructing the document image from text and layout modalities. With such generation capacity, \methodname{} is the first document AI model to achieve high-quality customizable, joint document editing and generation. 

Finally, our uniform sequence-to-sequence generation framework enables us to conveniently incorporate all major document supervised learning tasks to pretraining, i.e., document layout analysis, information extraction, document classification, document Q\&A, and Table QA/NLI, despite their significant differences in task and data format. In contrast, pretraining in previous document AI works is constrained to unlabeled data only (or using one single auxiliary supervised dataset such as FUNSD \cite{xu2021layoutlmv2}), while abundant labeled datasets with high quality supervision signals are ignored due to the lack of modeling flexibility. Overall, \methodname{} is pretrained on 11M public unlabeled documents, together with 11 supervised datasets of 1.8M examples. Ablation study in \Cref{tab:ablation} shows that \methodname{} only pretrained with the proposed self-supervised objectives exhibits great improvements over previous models, and adding the supervised data to pretraining further improves the performance.

We evaluate \methodname{} on \funsd{}~\cite{jaume2019funsd}, \cord{}~\cite{park2019cord}, \rvlcdip{}~\cite{harley2015evaluation}, \docvqa{}~\cite{mathew2021docvqa}, and \duebenchmark{}~\cite{borchmann2021due}. \methodname{} ranks the 1st place on the \duebenchmark{} leaderboard with 7 tasks, and also achieves SOTA on \cord{}, hence making \methodname{} a powerful and unified foundation Document AI model for diverse document understanding tasks,

To summarize, our major contributions include: 
    
    1. Unified representations and modeling for vision, text and layout modalities in document AI.
    
    2. Unified all document tasks to the sequence-to-sequence generation framework.
    
    3. Combined novel self-supervised objectives with supervised datasets in pretraining for unified document pretraining.
    
    4. \methodname{} can process and generate text, vision, and layout modalities together, which to the best of our knowledge is first one in the field of document AI.
    
    5. \methodname{} is a foundation model for Document AI, achieving SOTA on 8 tasks with significant margins.

\section{Related Work}

\noindent \textbf{Unifying Model Architectures in Multimodal Learning.} Unifying model architectures for different modalities, such as vision, language, and speech, is an emergent direction. Inspired by the immense success in natural language processing, computer vision and speech processing, model architectures in multimodal learning is converging to Transformers. One type of works concatenates text token embeddings and projected image patches as the input \cite{chen2020uniter,su2019vl} to a multimodal Transformer. Other models uses two-tower or three-tower architecture where each modality is encoded respectively. Projection heads or fusion networks on top of the two-tower architecture generate multimodal representations \cite{radford2021learning,yang2022code}.

\noindent \textbf{Unifying Tasks with the Generative Framework.} Research on unifying training processes across different tasks and domains recently has made significant progress. \cite{chung2022scaling} finetunes language models with instructions on 1.8k tasks. \cite{cho2021unifying} unifies several vision-language tasks by converting training objectives to sequence generation. \cite{wang2022ofa,wang2022image,lu2022unified} further combines more tasks, e.g., image generation, by converting images and bounding boxes to discrete tokens.

\noindent \textbf{Document Artificial Intelligence.} LayoutLM \cite{xu2020layoutlm} pretrains BERT models on document data with masked language modeling and document classification task, with 2D positional information and image embeddings integrated. Subsequent works \cite{xu2021layoutlmv2,huang2022layoutlmv3,hong2022bros} also adopt VL-BERT alike architecture and includes additional pretraining tasks, e.g., masked image/region modeling proposed, and leverages the reading order in layout information \cite{gu2022xylayoutlm}. \cite{gu2021unidoc,li2021selfdoc} use a multimodal encoder to model region features extracted by CNN with sentence-level text representations and train with self-supervised objectives. \cite{kim2022ocr} proposes an OCR-free model to directly generate textual output from document images. \cite{powalski2021going} trains generative language models on both unlabeled and labeled document data using generative training objectives. \cite{garncarek2021lambert} proposed to model documents as collections of tokens bounding boxes.

\section{Universal Document Processing}
\label{sec:method}
We introduce \methodname{}, a novel document AI framework with unified learning objectives and model architecture for text, vision, and layout as shown in \Cref{fig:architecture}. In this section, we will concretely discuss the proposed Vision-Text-Layout Transformer in \methodname{}, and will introduce the unified generative pretraining method in the next section. In document processing, given a document image $\vv$, typically optical character recognition (OCR) is used on $\vv$ to extract text tokens $\{s_i\}$ in the document and their bounding boxes $\{(x^1_i, y^1_i, x^2_i, y^2_i)\}$, i.e., the layout information for each token. $(x^1_i, y^1_i)$ and $(x^2_i, y^2_i)$ respectively represent the coordinates of the left-upper and right-bottom corner of the bounding box. Thus, suppose we have $M$ word tokens, the input is the triple, $(\vv, \{s_i\}^M_{i=1}, \{(x^1_i, y^1_i, x^2_i, y^2_i)\}^M_{i=1})$. \Cref{fig:architecture} shows an example document (left) and downstream tasks (right). 

\subsection{A Unified Vision, Text, and Layout Encoder}
\label{sec:encoder}
We fuse the vision, text, and layout modalities in the input stage using one unified transformer encoder. For traditional vision-text data, the text modality is usually the high-level description of the corresponding image or task prompt (e.g., question). While in document images, text is embedded inside the image, i.e., text and image pixels have one-to-one correspondence. To leverage this correspondence, we propose a new Vision-Text-Layout (VTL) Transformer architecture to dynamically fuse and unite the image pixels and text tokens based on the layout information.

Concretely, given the document image $\vv \in \mathbb{R}^{H \times W \times C}$, $M$ word tokens 
$\{s_i\}_{i=1}^M$ inside the image and the extracted layout structure $\{(x^1_i, y^1_i, x^2_i, y^2_i)\}_{i=1}^M$, we first partition $\vv$ into $\frac{H}{P}\times \frac{W}{P}$ image patches, where each patch is of size $P \times P\times C$. We then encode each patch with a $D$-dim vector and group all patch embeddings into a sequence of vectors $\{\vv_i \in \R^D\}_{i=1}^N$ where $N=\frac{H}{P}\times \frac{W}{P}$. Text tokens are also converted to numerical $D$-dim embeddings $\{\vs_i\}_{i=1}^M$ by vocabulary look-up.
\vspace{-10px}

\paragraph{Layout-Induced Vision-Text Embedding.}

\begin{figure}[t]
  \centering
  \includegraphics[width=\columnwidth]{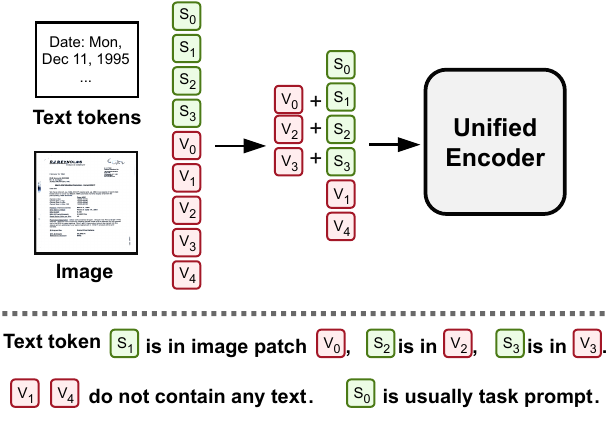}
  \caption{
    Layout-induced vision-text embedding.
}
\label{fig:vl_embed}
\vspace{-15px}
\end{figure}

Next, we build a unified representation for vision, text, and layout as shown in \Cref{fig:vl_embed}. We define the layout indicator function $\phi$ of image patch and token embeddings as follows:

\begin{equation}
\phi(\vs_i, \vv_j)=
\begin{cases}
   &1, \,\,\, \text{if the center of $\vs_i$'s bounding box} \\
   &\,\,\,\,\,\,\,\,\, \text{is within the image patch $\vv_j$.} \\
   &0, \,\,\, \text{otherwise}.
\end{cases}
\end{equation}
Then for each text token embedding $\vs_i$, the joint representation is the sum of its image patch feature\footnote{Some text token like manually crafted prompts have no locations. So, we set their layout bounding boxes to be $(0,0,0,0)$, i.e., they fall into a pseudo image patch.} and the text feature:
\vspace{-8px}
$$\vs'_i = \vs_i + \vv_j, \, \text{where} \,\, \phi(\vs_i, \vv_j) = 1.$$
\vspace{-10px}

For image patches $\vv_j$ without any text tokens, i.e. $\forall i, \, \phi(\vs_i, \vv_j) = 0$, the joint representation, $\vv'_j$ is itself: 
$$\vv'_j = \vv_j.$$ 
Note we do not have a designated joint representation for image patch containing tokens, since features of these image patches are already integrated with the text embeddings. Then $\{\vs_i'\}$ and $\{\vv_j'\}$ are fed into the VTL transformer encoder. These joint representations greatly enhance the interaction between vision, text and layout in the model input stage by explicitly leveraging their spatial correlations.

To further unify layout and text representation, inspired by the recent progress in generative object detection \cite{chen2022pixseq,wang2022ofa}, we discretize the layout modality, i.e., continuous coordinates text bounding box, to layout tokens. Suppose we have bounding box $(x^1_i, y^1_i, x^2_i, y^2_i)$ normalized in $[0,1]$. The resulting layout token will be each coordinate multiplied by vocabulary size and then rounded to nearest integer.
For example, if we have bounding box $(0.1, 0.2, 0.5, 0.6)$ with layout vocabulary size $500$, the layout tokens will then be \textit{<50><100><250><300>}. 
Layout tokens can be conveniently inserted into text context, and elegantly used for layout generation tasks (e.g., location detection). More details are discussed in \Cref{sec:pretrain}.

\begin{table*}[t]
\caption{A summary of all generative pretraining objectives with task names, task prompts, and task targets.}
\label{tab:pretraining_objectives}
\centering
\resizebox{1\textwidth}{!}{
\begin{tabular}{lp{0.46\textwidth}p{0.29\textwidth}}
\toprule
\textbf{Self-Supervised Tasks} & \textbf{Task Prompts} & \textbf{Task Targets} \\
\midrule
Layout Modeling & \textit{Layout Modeling}. {<layout\_0>} Ship Date to Retail {</layout\_0>} Week of March 14, 1994 & {<layout\_0>} \textit{<100><350><118><372>} \\ \hdashline
Visual Text Recognition & \textit{Visual Text Recognition}. {<text\_0>} \textit{<100><350><118> <372>} {</text\_0>} to Retail: Week of March 14, 1994   & {<text\_0>} Ship Date \\ \hdashline
Joint Text-Layout Reconstruction & \textit{Joint Text-Layout Reconstruction}. {<text\_layout\_0>} to Retail: Week of March 14, 1994 & {<text\_layout\_0>} Ship Date \textit{<100> <350><118><372>} \\ \hdashline
Masked Image Reconstruction & \textit{Masked Image Reconstruction.} Ship Date to Retail: Week of March 14, 1994  & [Pixels of the original image] \\
\toprule
\textbf{Supervised Tasks}\\ \toprule
Classification & \textit{Document Classification}. Ship Date to Retail: Week of March 14, 1994 & Memo. \\ \hdashline
Layout Analysis & \textit{Layout Analysis}. Paragraph. & Paragraph \textit{<82><35><150><439>} \\ \hdashline
Information Extraction & \textit{Information Extraction}. Ship Date to Retail & Week of March 14, 1994  \\ \hdashline
Question Answering & \textit{Question Answering}. What is the ship year? & 1994 \\ \hdashline
Document NLI & \textit{Document Natural Language Inference}. Ship Date to Retail: Week of March 14, 1994  & Entailment. \\
\bottomrule
\end{tabular}
}
\vspace{-4px}
\end{table*}

\paragraph{Position Bias.} We follow TILT~\cite{powalski2021going} to encode 2D text token position as 2D relative attention bias, similar to the relative attention bias used in T5. However, unlike T5, TILT, or transformer models in previous Document AI works \cite{huang2022layoutlmv3,powalski2021going}, we do not use 1D position embeddings in VTL transformer encoder, since the joint embedding and the 2D position bias already incorporate the layout structure of the input document.

\subsection{Vision-Text-Layout Decoder}
\label{sec:decoder}

As introduced in the previous section, the VTL encoder is able to compactly and jointly encode vision, text, and their layout. To perform various document generative tasks (will be discussed in \Cref{sec:pretrain}), the VTL decoder is designed to jointly generate all vision, text, and layout modalities.

The VTL decoder consists of a text-layout decoder and a vision decoder, as shown in \Cref{fig:architecture} (middle). The text-layout decoder is a uni-directional Transformer decoder to generate text and layout tokens in a sequence-to-sequence manner. For the vision decoder, we adopt the decoder of MAE \cite{he2021masked} and directly generate the image pixels with text and layout information. Details of the image decoding process will be discussed in the segment \textbf{``Masked Image Reconstruction with Text and Layout
''} of \Cref{sec:ss_tasks}. Both text-layout decoder and vision decoder will cross-attend to the VTL encoder.

Information such as model configurations are presented in \Cref{sec:modelconfig}.

\section{Unified Generative Pretraining}
\label{sec:pretrain}

To unify across different training objectives and datasets, we create a universal generative task format with task prompt. We pretrain \methodname{} on large-scale documents with and without human labels. We summarize the tasks prompts and targets in \Cref{tab:pretraining_objectives} which includes all self-supervised and supervised tasks respectively in upper and lower blocks. 

\subsection{Self-Supervised Pretraining Tasks}
\label{sec:ss_tasks}
We propose various innovative self-supervised learning objectives for unlabeled documents. The unlabeled document contains OCR text inputs with token-level bounding boxes and the document image. In the rest of this subsection, we use the following input text as example:\\
``{Ship Date to Retail: Week of March 14, 1994}''

\vspace{2px}

\noindent\textbf{(1) Joint Text-Layout Reconstruction} requires the model to reconstruct the missing texts and locate them in the document image. Concretely, we mask a percentage of text tokens and ask the model to both the tokens and their bounding boxes (i.e. layout tokens). E.g., assume masking ``Ship Date'' and ``of'', the input sequence and target sequence is given below:

\begin{table}[H]
\centering
\vspace{-8pt}
\resizebox{0.97\columnwidth}{!}{
\begin{tabular}{l}
\toprule
\textbf{Input Sequence:}\\
``\textit{Joint Text-Layout Reconstruction}. {<text\_layout\_0>}\\ 
to Retail: Week {<text\_layout\_1>} March 14, 1994'' \\
\midrule
\textbf{Target Sequence:}\\
``{<text\_layout\_0>} Ship Date \textit{<100><350><118><372>}\\ 
{<text\_layout\_1>} of \textit{<100><370><118><382>}'' \\
\bottomrule
\end{tabular}
}
\vspace{-10pt}
\end{table}

Here <text\_layout\_0> and <text\_layout\_1> denote the text-layout sentinel tokens, \textit{<100><350><118><372>} and \textit{<100><370><118><382>}'' represent the layout tokens of ``Date to'' and ``of'' respectively. We use masking ratio 15\% similar to Masked Language Modeling (MLM)~\cite{devlin2018bert} as this task can be interpreted as masked text-layout modeling. 

\noindent\textbf{(2) Layout Modeling}
asks the model to predict positions of (group of) text tokens, given the document image and context text. E.g., to predict positions of ``Ship Date'' and ``of'', the input sequence and target sequence is given below:

\begin{table}[H]
\centering
\vspace{-5pt}
\resizebox{0.97\columnwidth}{!}{
\begin{tabular}{l}
\toprule
\textbf{Input Sequence:}\\
``\textit{Layout Modeling}. {<layout\_0>} Ship Date {</layout\_0>} \\
to Retail: Week {<layout\_1>} of {</layout\_1>} March 14,\\ 1994'' \\
\midrule
\textbf{Target Sequence:}\\
``{<layout\_0>} \textit{<100><350><118><372>} {<layout\_1>}\\ \textit{<100><370><118><382>}'' \\
\bottomrule
\end{tabular}
}
\vspace{-5pt}
\end{table}

Note this pretraining task has a different sentinel token, <layout\_sent\_0>, from the previous task ``Joint Text-Layout Reconstruction'' because the generation content is different (layout vs. text + layout). We use large masking ratio 75\% since masking with small ratio results in an easy task.

\noindent\textbf{(3) Visual Text Recognition} identifies text at given location in the image. E.g., to recognize the text tokens at \textit{<100><350><118><372>} and \textit{<100><370><118><382>}, the input and target is:

\begin{table}[H]
\centering
\vspace{-5pt}
\resizebox{0.999\columnwidth}{!}{
\begin{tabular}{l}
\toprule
\textbf{Input Sequence:}\\
``\textit{Visual Text Recognition}. {<text\_0>} \textit{<100><350><118>}\\\textit{<372>} {</text\_0>} to Retail: Week {<text\_1>} \textit{<100><370>}\\\textit{<118><382>} {</text\_1>} March 14, 1994''
 \\
\midrule
\textbf{Target Sequence:}\\
``{<text\_0>} Ship Date {<text\_1>} of''  \\
\bottomrule
\end{tabular}
}
\vspace{-5pt}
\end{table}

Note this pretraining task also has a different sentinel token, <text\_0> .
We use masking ratio 50\% to distinguish this task from ``Joint Text-Layout Reconstruction'' and set the layout (bounding box) of sentinel token, e.g., <text\_0>, and layout token, e.g., \textit{<0><10><2><20>}, to (0,0,0,0). This objective helps model learn joint vision-text embedding by understanding vision-text correspondence.

\begin{figure}[t]
  \centering
  \includegraphics[width=1.0\columnwidth]{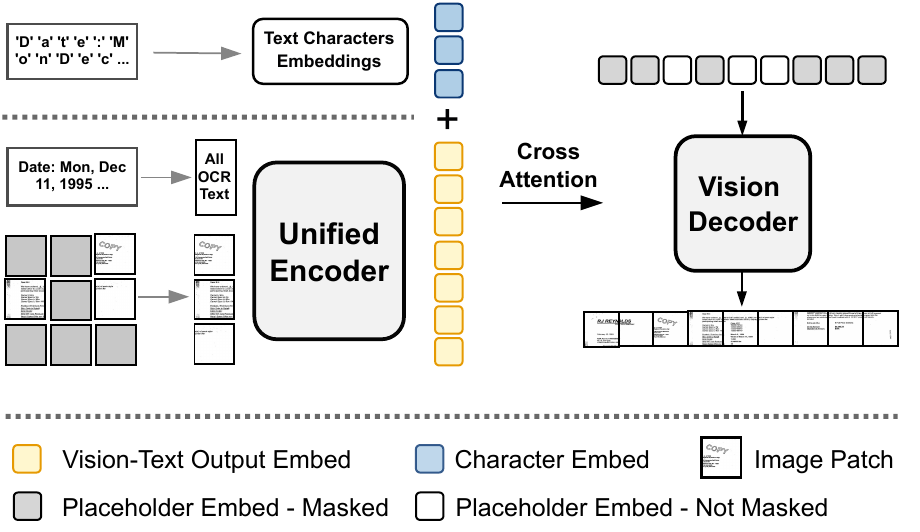}
  \caption{
    Masked autoencoding with text and layout.
}
\label{fig:mae}
\vspace{-2px}
\end{figure}

\noindent\textbf{(4) Masked Image Reconstruction with Text and Layout} aims to reconstruct image with text and layout as shown in \Cref{fig:mae}.
We adopt the MAE objective \cite{he2021masked} for vision self-supervised learning. Originally, MAE masks a percentage of the image patches and feed non-masked patches into a vision encoder. It then feeds encoder outputs to a vision decoder to reconstruct masked patches. MAE uses mean squared error and apply loss only on masked patches. We make the following modifications to the MAE decoding process to customize it for document image generation and our task unification framework:

\noindent\textbf{(4.a) Cross-Attention with Character Embeddings.} In document, the textual content mostly consists of alphabetic characters, numbers and punctuation. The character-level composition of text tokens should be helpful for the vision generation. We add cross-attention in the vision decoder that it attends to both the text token encoder features and embeddings of characters in the token (\Cref{fig:mae} left upper). These characters embeddings are trainable parameters and not encoded by the encoder. This cross-attention with characters only adds linear computation complexity but considerably improves the image generation quality.

\noindent\textbf{(4.b) Image Decoding.} Next, we describe the MAE decoding process.
For \methodname{}, we cannot directly feed the unified encoder output to the vision decoder, since the joint vision-text embedding only contains non-masked image patches to the unified encoder (\Cref{sec:encoder}), and image patches are fused with text tokens. Therefore, we propose that the vision decoder takes in a sequence of trainable placeholder embeddings. The length and order of the placeholder sequence is same as the patches of target image. We use two types of placeholder embeddings to indicate whether the image patch is masked in the input document image. The vision decoder attends to encoder vision-text output AND character embeddings via cross-attention. The above process is illustrated in \Cref{fig:mae}. We show the high quality generation visualization in \Cref{sec:vis_analysis}.

\subsection{Supervised Pretraining Tasks}
\label{sec:s_tasks}

Self-supervised tasks leverage large-scale unlabeled data to learn robust representations. On the other hand, supervised tasks use labeled data for fine-grained model supervision. We include the following supervised tasks in pretraining: document classification, layout analysis, information extraction, question answering, and document natural language inference. Details of the following supervised dataset are in \Cref{supp:ss_tasks}. Note that we do not conduct self-supervised tasks on the supervised datasets since we already have large-scale and diverse unlabeled data. Note that the validation or test set of downstream tasks is not used in supervised pretraining.

\noindent\textbf{Classification.} The task is to predict the document type. The task prompt is ``\textit{Document Classification on (Dataset Name)}'' like ``\textit{Document Classification on RVLCDIP}'', then followed by text tokens. The target is the document class. We use \rvlcdip{}~\cite{harley2015evaluation} with 16 document categories.

\noindent\textbf{Layout Analysis.}  This task is to predict locations of an entity in the document like title, paragraph, etc. The task prompt is ``\textit{Layout Analysis on (Dataset Name)}'', then followed by the entity name. The target are all bounding boxes that cover the given entity. We use \publaynet{}~\cite{zhong2019publaynet}.

\noindent\textbf{Information Extraction.} This task predict the entity type and location of a text query (e.g., the abstract paragraph). The task prompt is ``\textit{Information Extraction on (Dataset Name) (Text Query)}''. The target is the entity label and the bounding box of each token of the query.
We use \docbank{}~\cite{li2020docbank}, \kleister{} (KLC) ~\cite{stanislawek2021kleister}, \pwc{}~\cite{kardas2020axcell}, and \deepform{}~\cite{svetlichnaya2020deepform}.

\noindent\textbf{Question Answering.} The task is to answer a given question associated with the document image. The task prompt is ``\textit{Question Answering on (Dataset Name)}'', then followed by the question and all document tokens. The target is the answer. We use \websrc{}~\cite{chen2021websrc}, \visualmrc{}~\cite{tanaka2021visualmrc}, \docvqa{}~\cite{mathew2021docvqa}, \infovqa{}~\cite{mathew2022infographicvqa}, and \wtq{} (WikiTableQuestions)~\cite{pasupat2015compositional}.

\noindent\textbf{Document NLI.} Document Natural Language Inference predicts the entailment relationship between two sentences in a document. The prompt is ``\textit{Document Natural Language Inference on (Dataset Name)}'', then followed by the sentence pair. The target is the ``Entailment'' or ''Not Entailment''. We use \tabfact{}~\cite{chen2019tabfact} for this task.

\section{Experimental Setup}
\label{sec:expsetup}

\subsection{Model Pretraining}
\label{sec:modelconfig}
\noindent\textbf{Model Configuration.} In \methodname{}, the unified encoder and text-layout decoder follows the encoder-decoder architecture of T5-large \cite{raffel2020exploring}. The vision decoder is MAE-large decoder \cite{he2021masked}. Overall \methodname{} has 794M trainable parameters. For tokenizer, we use T5 tokenizer and embedding from Hugging Face Transformers \cite{wolf-etal-2020-transformers}. We also extend the vocabulary to accommodate special tokens (e.g., new sentinel and layout tokens).

\noindent\textbf{Data.} For self-supervised learning, we use \iitcdipfull{} \cite{lewis2006building}, a large-scale document collections commonly-used in previous works \cite{xu2020layoutlm,xu2021layoutlmv2,huang2022layoutlmv3}.
It contain 11 million scanned document with contains text and token-level bounding boxes extracted by OCR. Supervised datasets are as introduced in \Cref{sec:s_tasks}.

\noindent\textbf{Curriculum Learning.} We use large image resolution, $1024$, in our final settings since low resolution makes document text unidentifiable for both detection and generation. It will result in $(1024/16)^2=4096$ image patch sequence length which takes longer training time than small image resolution, e.g., $224$. Therefore, we use curriculum learning to start from a relatively small resolution and gradually scale up to 1024 resolution. In practice, we use scale with 3 resolutions during the pretraining $224\rightarrow512\rightarrow 1024$. We show the performance of the 3 stages in \Cref{supp:cl}.

\noindent\textbf{Training.} We use Adam~\cite{kingma2014adam} optimizer with learning rate 5e-5, 1000 warmup steps, batch size 512, weight decay of 1e-2, $\beta_1=0.9$, and $\beta_2=0.98$. For each curriculum learning stage, we train for 1 epoch.

\subsection{Downstream Evaluations}
\begin{table*}[t]
\caption{
Comparison with existing published models on the \duebenchmark{}. Modality T, L, V denote text, layout, or vision.}
\label{tab:due_results}
\centering
\resizebox{\textwidth}{!}{
\begin{tabular}{l c cc ccc cc c}
\toprule
\multirow{2}{*}{Model} & \multirow{2}{*}{Modality} & \multicolumn{2}{c}{Question Answering} &  \multicolumn{3}{c}{Information Extraction} &  \multicolumn{2}{c}{Table QA/NLI} & \multirow{2}{*}{Avg.} \\
\cmidrule(lr){3-4} \cmidrule(lr){5-7} \cmidrule(lr){8-9} 
& & \docvqa{} & InfoVQA & KLC & \pwc{} & \deepform{} & \wtq{} & \tabfact{} &\\
\midrule
Donut~\cite{kim2021donut} & V & 72.1 & - & - & - & - & - & - & - \\
BERT\textsubscript{large}~\cite{devlin2018bert} & T & 67.5 & - & - & - & - & - & - & - \\
T5\textsubscript{large}~\cite{raffel2020exploring} & T & 70.4 & 36.7 & 74.3 & 25.3 & 74.4 & 33.3 & 58.9 & 50.7 \\
T5\textsubscript{large}+U~\cite{powalski2021going} & T & 76.3 & 37.1 & 76.0 & 27.6 & 82.9 & 38.1 & 76.0 & 56.5 \\
T5\textsubscript{large}+2D~\cite{powalski2021going} & T+L & 69.8 & 39.2 & 72.6 & 25.7 & 74.0 & 30.8 & 58.0 & 50.4 \\
T5\textsubscript{large}+2D+U~\cite{powalski2021going} & T+L & 81.0 & 46.1 & 75.9 & 26.8 & 83.3 & 43.3 & 78.6 & 59.8 \\
LAMBERT~\cite{garncarek2021lambert} & T+L & - & - & 81.3 & - & - & - & - & - \\
StructuralLM\textsubscript{large}\cite{li2021structurallm} & T+L & 83.9 & - & - & - & - & - & - & - \\
LayoutLMv2\textsubscript{large}~\cite{xu2021layoutlmv2} & V+T+L & 78.8 & - & - & - & - & - & - & - \\
LayoutLMv3\textsubscript{large}~\cite{huang2022layoutlmv3} & V+T+L & 83.4 & 45.1 & 77.1 & 26.9 & 84.0 & 45.7 & 78.1 & 62.9 \\ 
\textbf{\methodname{}} & V+T+L & \textbf{84.7} & \textbf{47.4} & \textbf{82.8} & \textbf{28.0} & \textbf{85.5} & \textbf{47.2} & \textbf{78.9} & \textbf{64.8} \\
\bottomrule
\end{tabular}
}
\end{table*}

\begin{table}[t]
\caption{
Performance on \funsd{}, \cord{}, and \rvlcdip{} datasets. Modality V, T, L denote vision, text and layout.
}
\label{tab:main_results}
\centering
\resizebox{1.01\columnwidth}{!}{
\begin{tabular}{l c cccc}
\toprule
\multirow{2}{*}{Model} & \multirow{2}{*}{Modality} & \multicolumn{2}{c}{Info Ext.} &  \multicolumn{1}{c}{Classification} \\
\cmidrule(lr){3-4} \cmidrule(lr){5-5}
 & & \funsd{} & \cord{} & \rvlcdip{} \\
\midrule
Donut~\cite{kim2021donut} & V & - & 91.6 & 95.3 \\
BERT\textsubscript{large}~\cite{devlin2018bert} & T & 65.63 & 90.25 & 89.92 \\
BROS\textsubscript{large}~\cite{hong2022bros} & T+L & 84.52 & 97.40 & - \\
StructuralLM\textsubscript{large}~\cite{li2021structurallm} & T+L & 85.14 & - & \textbf{96.08} \\
LiLT\cite{wang2022lilt} & T+L & 88.41 & 96.07 & 95.68 \\
FormNet~\cite{lee2022formnet} & T+L & 84.69 & 97.28 & - \\
LayoutLM\textsubscript{large}~\cite{xu2020layoutlm} & T+L & 77.89 & - & 91.90 \\
SelfDoc~\cite{li2021selfdoc} & V+T+L & 83.36 & - & 92.81 \\
UniDoc~\cite{gu2021unidoc} & V+T+L & 87.93 & 96.86 & 95.05 \\
DocFormer\textsubscript{large}~\cite{appalaraju2021docformer} & V+T+L & 84.55 & 96.99 & 95.50 \\
TILT\textsubscript{large}~\cite{powalski2021going} & V+T+L & - & 96.33 & 95.52 \\
LayoutLMv2\textsubscript{large}~\cite{xu2021layoutlmv2} & V+T+L & 84.20 & 96.01 & 95.64 \\
LayoutLMv3\textsubscript{large}~\cite{huang2022layoutlmv3} & V+T+L & \textbf{92.08} & 97.46 & 95.93\\
\textbf{\methodname{}} & V+T+L & 91.62 & \textbf{97.58} & 96.00 \\
\bottomrule
\end{tabular}
}
\end{table}

We report the results on \funsd{}~\cite{jaume2019funsd}, \cord{}~\cite{park2019cord}, \rvlcdip{}~\cite{harley2015evaluation}, and \docvqa{}~\cite{mathew2021docvqa} in \Cref{tab:main_results} and describe their respective settings in below. We also report the results on 7 datasets of \duebenchmark{}~\cite{borchmann2021due} in \Cref{tab:due_results}. Finetuning training details are available in \Cref{sec:apd_finetune} and performance variance is available in \Cref{tab:var_due_results} and \Cref{tab:var_main_results}. Note that for all downstream tasks, we use the original OCR annotations provided in the datasets.

\textbf{\funsd{}} (Form Understanding in Noisy Scanned Documents \cite{jaume2019funsd}) has 149 and 50 samples for train and test. We evaluate on the entity recognition task: predicting the entity, "question", "answer", "header", or "other", for the text token. The task format is, suppose we have the title, "The Title", and its entity "[I-Header]", then the encoder input is "The Title" and the generation target is "The Title [I-Header]". The metric is F1 scores.

\textbf{CORD} (Consolidated Receipt Dataset for Post-OCR Parsing) ~\cite{park2019cord} is a key information extraction dataset with 30 labels under 4 categories such as "total" or "subtotal". It has 1,000 receipt samples. The train, validation, and test splits contain 800, 100, and 100 samples respectively. The metric is F1 and the task format is the same as FUNSD.

\textbf{\rvlcdip{}} is the document classification dataset that we have discussed previously. It has 320k/40k/40k images for training/validation/test. The metric is classification accuracy.

\textbf{\duebenchmark{}} contains 7 datasets and 3 domains, including document question answering (\docvqa{}~\cite{mathew2021docvqa}, InfographicsVQA\cite{mathew2022infographicvqa}), key information extraction (KLC\cite{stanislawek2021kleister}, \pwc{}\cite{kardas2020axcell}, \deepform{}\cite{svetlichnaya2020deepform}), and Table QA/NLI (\wtq{}\cite{pasupat2015compositional}, \tabfact{}\cite{chen2019tabfact}). Task prompt formats can be found in \Cref{sec:s_tasks} and details of datasets can be found in the appendix. 

\noindent\textbf{Results.}~Pretrained models are finetuned on each evaluation dataset. As shown in \Cref{tab:due_results}, our models \methodname{} 
achieve SOTA performance on all 7 tasks of \duebenchmark{}, ranking the 1st place on the leaderboard as of November 11, 2022. It also sets SOTA on \cord{} and (\Cref{tab:main_results}). It is worth noting that \methodname{} is an \textbf{open-vocabulary generative model} and uses \textbf{one single model for all tasks}. In comparison, most baselines leverage task-specific network for each dataset and are classification-based models. Nonetheless, \methodname{} still exhibits better results than those models.

Curriculum learning on image resolution (appendix \Cref{tab:cl_due_results}) shows that with larger resolution, \methodname{} steadily gains stronger performance. E.g., \methodname{} average performance on \duebenchmark{} with 224, 512 and 1024 resolution is 63.9, 64.3 and 65.1 respectively. Note our model with 224 resolution already outperform previous best models (e.g., average 62.9 on \duebenchmark{}). We then train \methodname{}
only with self-supervised objectives (224 resolution). Its performance (\Cref{tab:ablation}) also surpasses baselines, which shows the effectiveness of the unified representations, TVL transformer and the proposed self-supervised objectives. 

\begin{figure*}[t]
  \centering
  \includegraphics[width=1\textwidth]{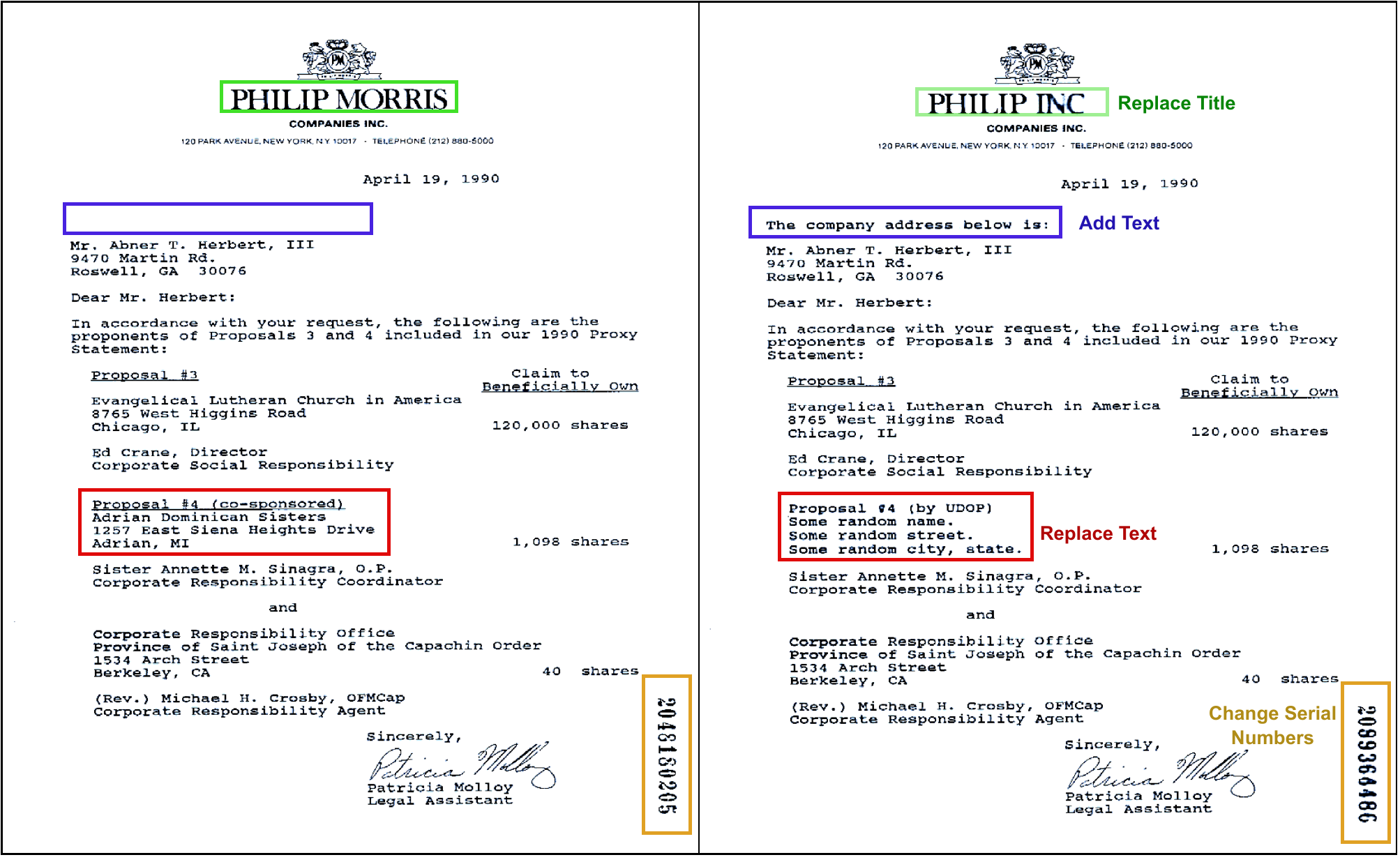}
  \caption{
  Document generation with customized content (right). Left is the original document. We show four document edits within the same figure including title replacement, text addition, text replacement, and tilted text replacement. All edits are done with one model run.
}
\label{fig:vis_mae_creative}
\end{figure*}

\begin{figure*}[t]
  \centering
  \includegraphics[width=1\textwidth]{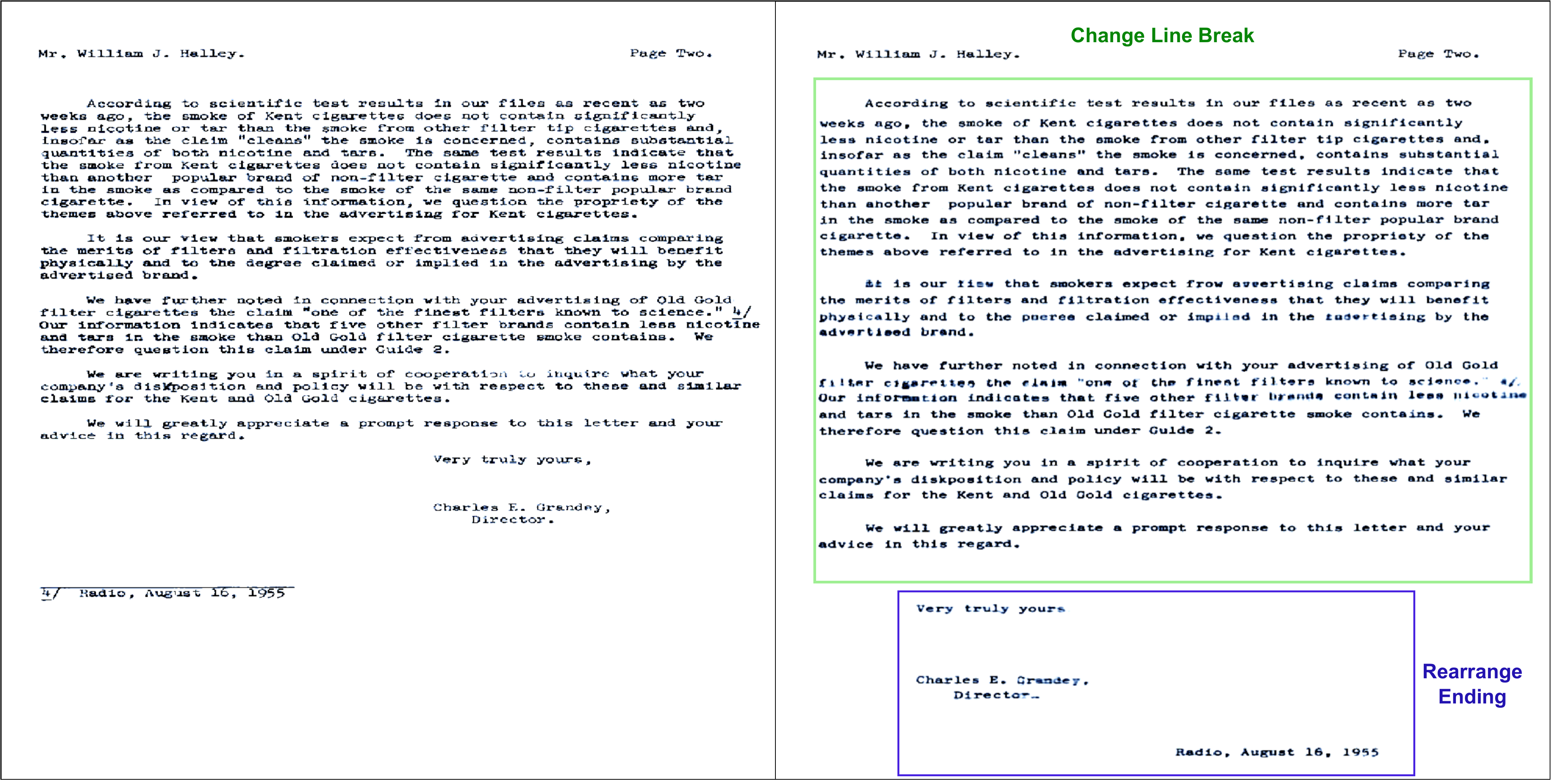}
  \caption{
  Document generation with customized layout (right). Left is the original document. We change the layout of the document text including line breaks change and text rearrangement. All edits are done with one model run.
}
\label{fig:vis_layout}
\end{figure*}

\section{Analysis}

\subsection{Visualization Analysis}
\label{sec:vis_analysis}
\begin{figure}[h]
  \centering
  \includegraphics[width=0.99\columnwidth]{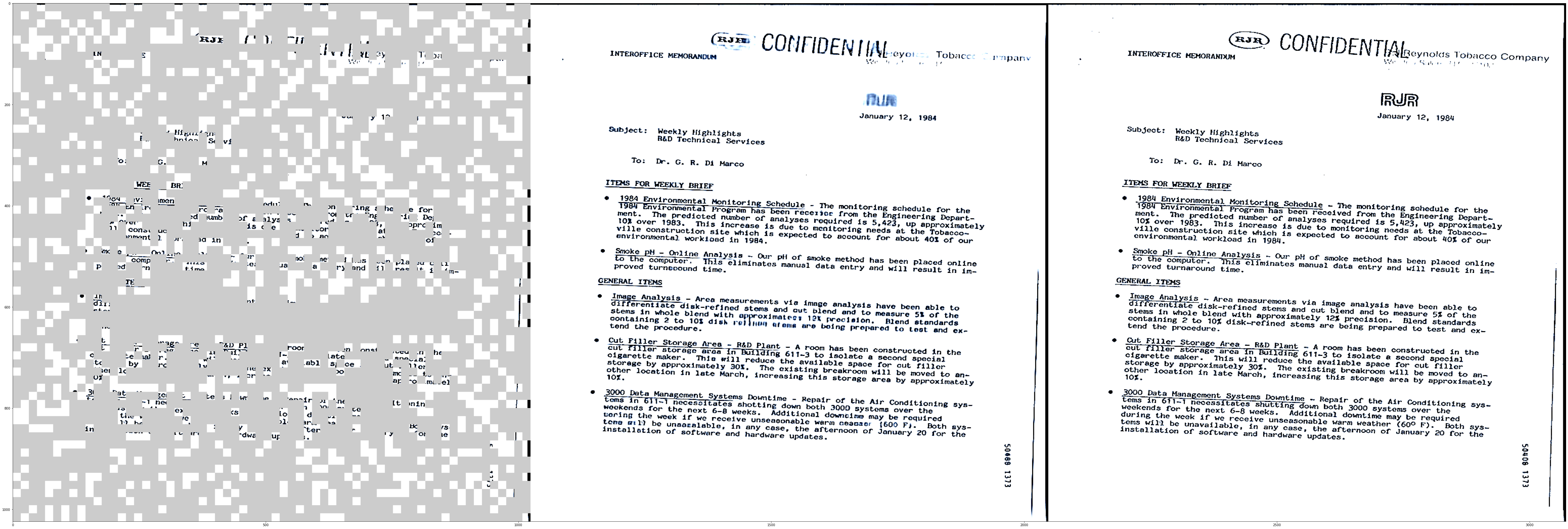}
  \includegraphics[width=0.99\columnwidth]{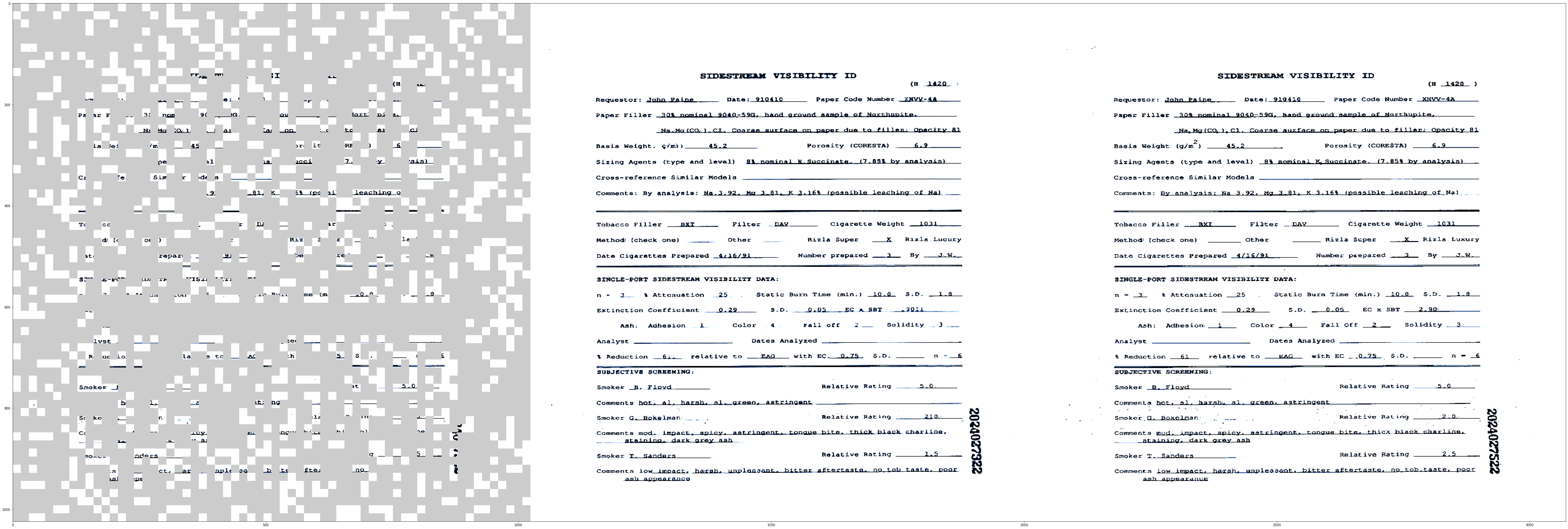}
  \caption{
    MAE demonstrations with 75\% masking. Middle: reconstruction, Right: original.
}
\label{fig:vis_mae_random}
\vspace{-8px}
\end{figure}

 \noindent \textbf{Masked Image Reconstruction.} \Cref{fig:vis_mae_random} presents masked image reconstruction. Even with high masking ratio, the model can reconstruct the document image from text and layout signals with high quality: reconstructed contents are clear, consistent, and almost identical with the original image (all demonstrations are conducted on unseen documents.).

\noindent \textbf{Document Generation \& Editing.} For the first time in Document AI, \methodname{} achieves controllable high-quality document generation and editing. As shown in \cref{fig:vis_mae_creative}), one can edit and add to the document image content with customized contents. The generated content is of high resolution and is consistent with the context in font, size, style and orientation (e.g., vertical numbers in \cref{fig:vis_mae_creative}). More generation examples are available in \Cref{supp:demos}. This is done by masking the regions to edit in the document image, and specifying the customized content in the text input, and their positions through layout embeddings. This novel functionality can generate augmentation document data for future research.

\noindent \textbf{Layout Customization.} \methodname{} can perform controllable high-quality document layout edits. We show examples in \Cref{fig:vis_layout}, where our model can edit the layout of the document by regenerating the document from scratch. This is done by keeping only a few image patch as prompt, change the bounding boxes of the content, and then regenerate the document image with the new layout.

\subsection{Ablation Analysis}
\label{sec:ablation_analysis}
\paragraph{Pretraining Objectives.}
\Cref{tab:ablation} presents the ablation study of pretraining objectives on \docvqa{} and \rvlcdip{} validation sets. We first develop a MLM (Masked Language Modeling) baseline that is a \methodname{} model pre-trained only on the BERT’s
MLM~\cite{devlin2018bert} that masks 15\% of the input tokens. \methodname{} models (224 image resolution) 
pretrained with layout/text self-supervised objectives (``Layout Modeling'', ``Visual Text Dataition'', and ``Joint Text-Layout Reconstruction'') outperforms the one trained with masked language modeling (MLM), confirming their effectiveness. \Cref{tab:ablation} also shows relative effectiveness of each pretraining task. Layout modeling improves upon Joint Text-Layout Modeling; Masked Image Reconstruction improves on text-based pretraining tasks. Adding vision self-supervised learning (masked image reconstruction) and supervised learning further improves the performance.

\begin{table}[h]
\caption{
Ablation study on pre-training objectives.}
\label{tab:ablation}
\centering
\resizebox{\columnwidth}{!}{
\begin{tabular}{l ccc c}
\toprule
Pretrain Objectives & \#Pretrain Data  & DocVQA & RVL-CDIP \\
\midrule
MLM & 11.0M & 79.7 $\pm$ 0.4 & 95.3 $\pm$ 0.3 \\
\midrule
Joint Text-Layout & 11.0M & 82.8 $\pm$ 0.1 & 95.4 $\pm$ 0.3\\
\hspace*{1em} + Visual Text Recognition & 11.0M & 83.3 $\pm$ 0.2 & 95.4 $\pm$ 0.2 \\
\hspace*{2em} + Layout Modeling & 11.0M & 84.0 $\pm$ 0.3 & 95.6 $\pm$ 0.2\\
\hspace*{3em} + Image Reconstruction & 11.0M & 84.4 $\pm$ 0.2 & 96.2 $\pm$ 0.2\\
\hspace*{4em} + Supervised & 12.8M & \textbf{85.0}$\pm$ 0.2  & \textbf{96.3} $\pm$ 0.1\\
\bottomrule
\end{tabular}
}
\end{table}

\begin{table*}[t]
\caption{
Ablations on model architecture.}
\label{tab:ablation_architecture}
\centering
\resizebox{0.7\textwidth}{!}{
\begin{tabular}{l c cc ccc cc c}
\toprule
\multirow{2}{*}{Model}  & \multicolumn{2}{c}{Question Answering} &  \multicolumn{3}{c}{Information Extraction} &  \multicolumn{2}{c}{Table QA/NLI} & \multirow{2}{*}{Avg.} \\
\cmidrule(lr){3-4} \cmidrule(lr){5-7} \cmidrule(lr){8-9} 
& & \docvqa{} & InfoVQA & KLC & \pwc{} & \deepform{} & \wtq{} & \tabfact{} &\\
\midrule
\textbf{\dual{}} & 84.4 & 47.1 & 81.9 & 28.0 & 85.2 & 46.7 & \textbf{79.5} & 64.6\\
\textbf{\methodname{}} & \textbf{84.7} & \textbf{47.4} & \textbf{82.8} & \textbf{28.0} & \textbf{85.5} & \textbf{47.2} & 78.9 & \textbf{64.8} \\
\bottomrule
\end{tabular}
}

\end{table*}

\paragraph{Modality-Specific Model Variant.}
In the field of multimodal learning, a common model architecture is the two-tower model, where vision and text are encoded by two modality-specific encoders respectively \cite{radford2021learning,yang2022code}. Therefore, we explore an variant of \methodname{} such that instead of having one unified encoder, we separately use a text encoder (to encode both text and layout tokens) and a vision encoder. Position bias are used in both encoders to represent layout information following previous works. We name this variant \dual{}. For \dual{}, the text-layout encoder-decoder follows T5-large, and the vision encoder-decoder has the same configuration as MAE-large. It has in total 1098M trainable parameters. As shown in \Cref{tab:ablation_architecture} and \Cref{tab:udop_apped}, using one unified encoder is better than having separated encoders in most datasets. The exceptions are WTQ and \rvlcdip{} on which \dual{} achieves SOTA.

\paragraph{Additional Supervised Training Stage}
\label{supp:additions}
TILT~\cite{powalski2021going} performs additional training on a wide range of QA datasets, such as reading comprehension dataset SQuAD \cite{rajpurkar2016squad}, before the finetuning on \docvqa{}. This results in considerable performance improvement of the TILT model on \docvqa{} and \infovqa{}. To have a fair comparison, we also finetune \methodname{} on the same set of datasets before testing on \docvqa{} or \infovqa{}. As shown in \Cref{tab:tilt_docvqa}, \methodname{} is further improved with this auxiliary training and outperforms TILT.

\begin{table}[h]
\caption{
Training \methodname{} on auxiliary QA datasets as in TILT. The performance of \methodname{} on \docvqa{} and \infovqa{} is further improved (performance without the auxiliary training was not reported in the TILT paper).
}
\label{tab:tilt_docvqa}
\centering
\resizebox{\columnwidth}{!}{
\begin{tabular}{l ccc}
\toprule
Model & DocVQA & InfoVQA \\
\midrule
TILT\textsubscript{large}(w/ auxiliary training) & 87.1 & 61.2 \\
UDOP (w/o auxiliary training) & 84.7 & 47.4 \\
UDOP (w/ auxiliary training) & \textbf{87.8} & \textbf{63.0}\\
\bottomrule
\end{tabular}
}
\end{table}

\subsection{Effectiveness of the Vision Modality} In the field of Document AI, the effectiveness of the vision modality, i.e., document images, is unclear. We explore this by removing the visual embedding from the model input, with results shown in \Cref{tab:image_embed}. It shows that the vision modality is more prominent on visually-rich tasks, e.g., \infovqa{}, compared with text-dominant data such as DocVQA.
\begin{table}[h]
\caption{
Effectiveness of the vision modality.
}
\label{tab:image_embed}
\centering
\resizebox{\columnwidth}{!}{
\begin{tabular}{l cc}
\toprule
Model & DocVQA & InfoVQA \\
\midrule
UDOP & \textbf{84.7} & \textbf{47.4} \\
UDOP w/o image input embeddings & 84.4 & 45.0 \\
\bottomrule
\end{tabular}
}
\end{table}

\section{Conclusion}
\label{sec:conclusion}
In this work, we propose \methodname{}, a foundation model for document AI. \methodname{} unifies the vision, text and layout modalities of documents by utilizing their strong spatial correlations through layout-induced vision-text representations and Vision-Text-Layout transformer. It also unites all self-supervised and supervised document tasks with a generative framework. \methodname{} achieves SOTA on 8 tasks and currently ranks the 1st place on the Document Understanding Benchmark Leaderboard. For the first time in document AI, \methodname{} achieves customizable realistic document generation and editing. We discuss the limitations and societal impact of our work in the appendix.

{\small
\bibliographystyle{ieee_fullname}
\bibliography{references}

\begin{thebibliography}{10}\itemsep=-1pt

\bibitem{appalaraju2021docformer}
Srikar Appalaraju, Bhavan Jasani, Bhargava~Urala Kota, Yusheng Xie, and R
  Manmatha.
\newblock Docformer: End-to-end transformer for document understanding.
\newblock In {\em Proceedings of the IEEE/CVF International Conference on
  Computer Vision}, pages 993--1003, 2021.

\bibitem{borchmann2021due}
{\L}ukasz Borchmann, Micha{\l} Pietruszka, Tomasz Stanislawek, Dawid
  Jurkiewicz, Micha{\l} Turski, Karolina Szyndler, and Filip Grali{\'n}ski.
\newblock Due: End-to-end document understanding benchmark.
\newblock In {\em Thirty-fifth Conference on Neural Information Processing
  Systems Datasets and Benchmarks Track (Round 2)}, 2021.

\bibitem{chen2021websrc}
Lu Chen, Xingyu Chen, Zihan Zhao, Danyang Zhang, Jiabao Ji, Ao Luo, Yuxuan
  Xiong, and Kai Yu.
\newblock Websrc: A dataset for web-based structural reading comprehension.
\newblock {\em arXiv preprint arXiv:2101.09465}, 2021.

\bibitem{chen2022pixseq}
Ting Chen, Saurabh Saxena, Lala Li, David~J. Fleet, and Geoffrey Hinton.
\newblock Pix2seq: A language modeling framework for object detection.
\newblock In {\em International Conference on Learning Representations}, 2022.

\bibitem{chen2019tabfact}
Wenhu Chen, Hongmin Wang, Jianshu Chen, Yunkai Zhang, Hong Wang, Shiyang Li,
  Xiyou Zhou, and William~Yang Wang.
\newblock Tabfact: A large-scale dataset for table-based fact verification.
\newblock {\em arXiv preprint arXiv:1909.02164}, 2019.

\bibitem{chen2020uniter}
Yen-Chun Chen, Linjie Li, Licheng Yu, Ahmed El~Kholy, Faisal Ahmed, Zhe Gan, Yu
  Cheng, and Jingjing Liu.
\newblock Uniter: Universal image-text representation learning.
\newblock In {\em European conference on computer vision}, pages 104--120.
  Springer, 2020.

\bibitem{cho2021unifying}
Jaemin Cho, Jie Lei, Hao Tan, and Mohit Bansal.
\newblock Unifying vision-and-language tasks via text generation.
\newblock In {\em International Conference on Machine Learning}, pages
  1931--1942. PMLR, 2021.

\bibitem{chung2022scaling}
Hyung~Won Chung, Le Hou, Shayne Longpre, Barret Zoph, Yi Tay, William Fedus,
  Eric Li, Xuezhi Wang, Mostafa Dehghani, Siddhartha Brahma, et~al.
\newblock Scaling instruction-finetuned language models.
\newblock {\em arXiv preprint arXiv:2210.11416}, 2022.

\bibitem{devlin2018bert}
Jacob Devlin, Ming-Wei Chang, Kenton Lee, and Kristina Toutanova.
\newblock Bert: Pre-training of deep bidirectional transformers for language
  understanding.
\newblock In {\em NAACL}, 2018.

\bibitem{garncarek2021lambert}
{\L}ukasz Garncarek, Rafa{\l} Powalski, Tomasz Stanis{\l}awek, Bartosz
  Topolski, Piotr Halama, Micha{\l} Turski, and Filip Grali{\'n}ski.
\newblock Lambert: layout-aware language modeling for information extraction.
\newblock In {\em International Conference on Document Analysis and
  Recognition}, pages 532--547. Springer, 2021.

\bibitem{gu2021unidoc}
Jiuxiang Gu, Jason Kuen, Vlad~I Morariu, Handong Zhao, Rajiv Jain, Nikolaos
  Barmpalios, Ani Nenkova, and Tong Sun.
\newblock Unidoc: Unified pretraining framework for document understanding.
\newblock {\em Advances in Neural Information Processing Systems}, 34:39--50,
  2021.

\bibitem{gu2022xylayoutlm}
Zhangxuan Gu, Changhua Meng, Ke Wang, Jun Lan, Weiqiang Wang, Ming Gu, and
  Liqing Zhang.
\newblock Xylayoutlm: Towards layout-aware multimodal networks for
  visually-rich document understanding.
\newblock In {\em Proceedings of the IEEE/CVF Conference on Computer Vision and
  Pattern Recognition}, pages 4583--4592, 2022.

\bibitem{harley2015evaluation}
Adam~W Harley, Alex Ufkes, and Konstantinos~G Derpanis.
\newblock Evaluation of deep convolutional nets for document image
  classification and retrieval.
\newblock In {\em 2015 13th International Conference on Document Analysis and
  Recognition (ICDAR)}, pages 991--995. IEEE, 2015.

\bibitem{he2021masked}
Kaiming He, Xinlei Chen, Saining Xie, Yanghao Li, Piotr Doll{\'a}r, and Ross
  Girshick.
\newblock Masked autoencoders are scalable vision learners.
\newblock {\em arXiv preprint arXiv:2111.06377}, 2021.

\bibitem{hong2022bros}
Teakgyu Hong, DongHyun Kim, Mingi Ji, Wonseok Hwang, Daehyun Nam, and Sungrae
  Park.
\newblock Bros: A pre-trained language model focusing on text and layout for
  better key information extraction from documents.
\newblock {\em Proceedings of the AAAI Conference on Artificial Intelligence},
  36(10):10767--10775, Jun. 2022.

\bibitem{huang2022layoutlmv3}
Yupan Huang, Tengchao Lv, Lei Cui, Yutong Lu, and Furu Wei.
\newblock Layoutlmv3: Pre-training for document ai with unified text and image
  masking.
\newblock {\em arXiv preprint arXiv:2204.08387}, 2022.

\bibitem{huang2020pixel}
Zhicheng Huang, Zhaoyang Zeng, Bei Liu, Dongmei Fu, and Jianlong Fu.
\newblock Pixel-bert: Aligning image pixels with text by deep multi-modal
  transformers.
\newblock {\em arXiv preprint arXiv:2004.00849}, 2020.

\bibitem{jaume2019funsd}
Guillaume Jaume, Hazim~Kemal Ekenel, and Jean-Philippe Thiran.
\newblock Funsd: A dataset for form understanding in noisy scanned documents.
\newblock In {\em 2019 International Conference on Document Analysis and
  Recognition Workshops (ICDARW)}, volume~2, pages 1--6. IEEE, 2019.

\bibitem{kardas2020axcell}
Marcin Kardas, Piotr Czapla, Pontus Stenetorp, Sebastian Ruder, Sebastian
  Riedel, Ross Taylor, and Robert Stojnic.
\newblock Axcell: Automatic extraction of results from machine learning papers.
\newblock {\em arXiv preprint arXiv:2004.14356}, 2020.

\bibitem{kim2022ocr}
Geewook Kim, Teakgyu Hong, Moonbin Yim, JeongYeon Nam, Jinyoung Park, Jinyeong
  Yim, Wonseok Hwang, Sangdoo Yun, Dongyoon Han, and Seunghyun Park.
\newblock Ocr-free document understanding transformer.
\newblock In {\em European Conference on Computer Vision}, pages 498--517.
  Springer, 2022.

\bibitem{kim2021donut}
Geewook Kim, Teakgyu Hong, Moonbin Yim, Jinyoung Park, Jinyeong Yim, Wonseok
  Hwang, Sangdoo Yun, Dongyoon Han, and Seunghyun Park.
\newblock Donut: Document understanding transformer without ocr.
\newblock {\em arXiv preprint arXiv:2111.15664}, 2021.

\bibitem{Kim2021ViLT}
Wonjae Kim, Bokyung Son, and Ildoo Kim.
\newblock {ViLT: Vision-and-Language Transformer Without Convolution or Region
  Supervision}.
\newblock In {\em ICML}, 2021.

\bibitem{kingma2014adam}
Diederik~P Kingma and Jimmy Ba.
\newblock Adam: A method for stochastic optimization.
\newblock In {\em ICLR}, 2014.

\bibitem{lee2022formnet}
Chen-Yu Lee, Chun-Liang Li, Timothy Dozat, Vincent Perot, Guolong Su, Nan Hua,
  Joshua Ainslie, Renshen Wang, Yasuhisa Fujii, and Tomas Pfister.
\newblock Formnet: Structural encoding beyond sequential modeling in form
  document information extraction.
\newblock {\em arXiv preprint arXiv:2203.08411}, 2022.

\bibitem{lewis2006building}
David Lewis, Gady Agam, Shlomo Argamon, Ophir Frieder, David Grossman, and
  Jefferson Heard.
\newblock Building a test collection for complex document information
  processing.
\newblock In {\em Proceedings of the 29th annual international ACM SIGIR
  conference on Research and development in information retrieval}, pages
  665--666, 2006.

\bibitem{li2021structurallm}
Chenliang Li, Bin Bi, Ming Yan, Wei Wang, Songfang Huang, Fei Huang, and Luo
  Si.
\newblock Structurallm: Structural pre-training for form understanding.
\newblock {\em arXiv preprint arXiv:2105.11210}, 2021.

\bibitem{li2019visualbert}
Liunian~Harold Li, Mark Yatskar, Da Yin, Cho-Jui Hsieh, and Kai-Wei Chang.
\newblock Visualbert: A simple and performant baseline for vision and language.
\newblock {\em arXiv preprint arXiv:1908.03557}, 2019.

\bibitem{li2020docbank}
Minghao Li, Yiheng Xu, Lei Cui, Shaohan Huang, Furu Wei, Zhoujun Li, and Ming
  Zhou.
\newblock Docbank: A benchmark dataset for document layout analysis.
\newblock {\em arXiv preprint arXiv:2006.01038}, 2020.

\bibitem{li2021selfdoc}
Peizhao Li, Jiuxiang Gu, Jason Kuen, Vlad~I Morariu, Handong Zhao, Rajiv Jain,
  Varun Manjunatha, and Hongfu Liu.
\newblock Selfdoc: Self-supervised document representation learning.
\newblock In {\em Proceedings of the IEEE/CVF Conference on Computer Vision and
  Pattern Recognition}, pages 5652--5660, 2021.

\bibitem{li2021structext}
Yulin Li, Yuxi Qian, Yuechen Yu, Xiameng Qin, Chengquan Zhang, Yan Liu, Kun
  Yao, Junyu Han, Jingtuo Liu, and Errui Ding.
\newblock Structext: Structured text understanding with multi-modal
  transformers.
\newblock In {\em Proceedings of the 29th ACM International Conference on
  Multimedia}, pages 1912--1920, 2021.

\bibitem{lu2022unified}
Jiasen Lu, Christopher Clark, Rowan Zellers, Roozbeh Mottaghi, and Aniruddha
  Kembhavi.
\newblock Unified-io: A unified model for vision, language, and multi-modal
  tasks.
\newblock {\em arXiv preprint arXiv:2206.08916}, 2022.

\bibitem{mathew2022infographicvqa}
Minesh Mathew, Viraj Bagal, Rub{\`e}n Tito, Dimosthenis Karatzas, Ernest
  Valveny, and CV Jawahar.
\newblock Infographicvqa.
\newblock In {\em Proceedings of the IEEE/CVF Winter Conference on Applications
  of Computer Vision}, pages 1697--1706, 2022.

\bibitem{mathew2021docvqa}
Minesh Mathew, Dimosthenis Karatzas, and CV Jawahar.
\newblock Docvqa: A dataset for vqa on document images.
\newblock In {\em Proceedings of the IEEE/CVF winter conference on applications
  of computer vision}, pages 2200--2209, 2021.

\bibitem{park2019cord}
Seunghyun Park, Seung Shin, Bado Lee, Junyeop Lee, Jaeheung Surh, Minjoon Seo,
  and Hwalsuk Lee.
\newblock Cord: a consolidated receipt dataset for post-ocr parsing.
\newblock In {\em Workshop on Document Intelligence at NeurIPS 2019}, 2019.

\bibitem{pasupat2015compositional}
Panupong Pasupat and Percy Liang.
\newblock Compositional semantic parsing on semi-structured tables.
\newblock {\em arXiv preprint arXiv:1508.00305}, 2015.

\bibitem{powalski2021going}
Rafa{\l} Powalski, {\L}ukasz Borchmann, Dawid Jurkiewicz, Tomasz Dwojak,
  Micha{\l} Pietruszka, and Gabriela Pa{\l}ka.
\newblock Going full-tilt boogie on document understanding with
  text-image-layout transformer.
\newblock In {\em International Conference on Document Analysis and
  Recognition}, pages 732--747. Springer, 2021.

\bibitem{pramanik2020towards}
Subhojeet Pramanik, Shashank Mujumdar, and Hima Patel.
\newblock Towards a multi-modal, multi-task learning based pre-training
  framework for document representation learning.
\newblock {\em arXiv preprint arXiv:2009.14457}, 2020.

\bibitem{radford2021learning}
Alec Radford, Jong~Wook Kim, Chris Hallacy, Aditya Ramesh, Gabriel Goh,
  Sandhini Agarwal, Girish Sastry, Amanda Askell, Pamela Mishkin, Jack Clark,
  et~al.
\newblock Learning transferable visual models from natural language
  supervision.
\newblock In {\em International Conference on Machine Learning}, pages
  8748--8763. PMLR, 2021.

\bibitem{raffel2020exploring}
Colin Raffel, Noam Shazeer, Adam Roberts, Katherine Lee, Sharan Narang, Michael
  Matena, Yanqi Zhou, Wei Li, and Peter~J Liu.
\newblock Exploring the limits of transfer learning with a unified text-to-text
  transformer.
\newblock {\em JMLR}, 2020.

\bibitem{rajpurkar2016squad}
Pranav Rajpurkar, Jian Zhang, Konstantin Lopyrev, and Percy Liang.
\newblock Squad: 100,000+ questions for machine comprehension of text.
\newblock In {\em EMNLP}, 2016.

\bibitem{stanislawek2021kleister}
Tomasz Stanis{\l}awek, Filip Grali{\'n}ski, Anna Wr{\'o}blewska, Dawid
  Lipi{\'n}ski, Agnieszka Kaliska, Paulina Rosalska, Bartosz Topolski, and
  Przemys{\l}aw Biecek.
\newblock Kleister: key information extraction datasets involving long
  documents with complex layouts.
\newblock In {\em International Conference on Document Analysis and
  Recognition}, pages 564--579. Springer, 2021.

\bibitem{su2019vl}
Weijie Su, Xizhou Zhu, Yue Cao, Bin Li, Lewei Lu, Furu Wei, and Jifeng Dai.
\newblock Vl-bert: Pre-training of generic visual-linguistic representations.
\newblock In {\em International Conference on Learning Representations}, 2019.

\bibitem{svetlichnaya2020deepform}
S Svetlichnaya.
\newblock Deepform: Understand structured documents at scale, 2020.

\bibitem{tan2019lxmert}
Hao Tan and Mohit Bansal.
\newblock Lxmert: Learning cross-modality encoder representations from
  transformers.
\newblock In {\em EMNLP}, 2019.

\bibitem{tanaka2021visualmrc}
Ryota Tanaka, Kyosuke Nishida, and Sen Yoshida.
\newblock Visualmrc: Machine reading comprehension on document images.
\newblock {\em Proceedings of the AAAI Conference on Artificial Intelligence},
  35(15):13878--13888, May 2021.

\bibitem{tang2022perceiver}
Zineng Tang, Jaemin Cho, Jie Lei, and Mohit Bansal.
\newblock Perceiver-vl: Efficient vision-and-language modeling with iterative
  latent attention.
\newblock {\em arXiv preprint arXiv:2211.11701}, 2022.

\bibitem{tang2021decembert}
Zineng Tang, Jie Lei, and Mohit Bansal.
\newblock Decembert: Learning from noisy instructional videos via dense
  captions and entropy minimization.
\newblock In {\em Proceedings of the 2021 Conference of the North American
  Chapter of the Association for Computational Linguistics: Human Language
  Technologies}, pages 2415--2426, 2021.

\bibitem{wang2022lilt}
Jiapeng Wang, Lianwen Jin, and Kai Ding.
\newblock Lilt: A simple yet effective language-independent layout transformer
  for structured document understanding.
\newblock In {\em Proceedings of the 60th Annual Meeting of the Association for
  Computational Linguistics (Volume 1: Long Papers)}, pages 7747--7757, 2022.

\bibitem{wang2022ofa}
Peng Wang, An Yang, Rui Men, Junyang Lin, Shuai Bai, Zhikang Li, Jianxin Ma,
  Chang Zhou, Jingren Zhou, and Hongxia Yang.
\newblock Ofa: Unifying architectures, tasks, and modalities through a simple
  sequence-to-sequence learning framework.
\newblock In {\em International Conference on Machine Learning}, pages
  23318--23340. PMLR, 2022.

\bibitem{wang2022image}
Wenhui Wang, Hangbo Bao, Li Dong, Johan Bjorck, Zhiliang Peng, Qiang Liu, Kriti
  Aggarwal, Owais~Khan Mohammed, Saksham Singhal, Subhojit Som, et~al.
\newblock Image as a foreign language: Beit pretraining for all vision and
  vision-language tasks.
\newblock {\em arXiv preprint arXiv:2208.10442}, 2022.

\bibitem{wolf-etal-2020-transformers}
Thomas Wolf, Lysandre Debut, Victor Sanh, Julien Chaumond, Clement Delangue,
  Anthony Moi, Pierric Cistac, Tim Rault, Rémi Louf, Morgan Funtowicz, Joe
  Davison, Sam Shleifer, Patrick von Platen, Clara Ma, Yacine Jernite, Julien
  Plu, Canwen Xu, Teven~Le Scao, Sylvain Gugger, Mariama Drame, Quentin Lhoest,
  and Alexander~M. Rush.
\newblock Transformers: State-of-the-art natural language processing.
\newblock In {\em Proceedings of the 2020 Conference on Empirical Methods in
  Natural Language Processing: System Demonstrations}, pages 38--45, Online,
  Oct. 2020. Association for Computational Linguistics.

\bibitem{wu2021lampret}
Te-Lin Wu, Cheng Li, Mingyang Zhang, Tao Chen, Spurthi~Amba Hombaiah, and
  Michael Bendersky.
\newblock Lampret: Layout-aware multimodal pretraining for document
  understanding.
\newblock {\em arXiv preprint arXiv:2104.08405}, 2021.

\bibitem{xu2020layoutlm}
Yiheng Xu, Minghao Li, Lei Cui, Shaohan Huang, Furu Wei, and Ming Zhou.
\newblock Layoutlm: Pre-training of text and layout for document image
  understanding.
\newblock In {\em Proceedings of the 26th ACM SIGKDD International Conference
  on Knowledge Discovery \& Data Mining}, pages 1192--1200, 2020.

\bibitem{xu2021layoutxlm}
Yiheng Xu, Tengchao Lv, Lei Cui, Guoxin Wang, Yijuan Lu, Dinei Florencio, Cha
  Zhang, and Furu Wei.
\newblock Layoutxlm: Multimodal pre-training for multilingual visually-rich
  document understanding.
\newblock {\em arXiv preprint arXiv:2104.08836}, 2021.

\bibitem{xu2021layoutlmv2}
Yang Xu, Yiheng Xu, Tengchao Lv, Lei Cui, Furu Wei, Guoxin Wang, Yijuan Lu,
  Dinei Florencio, Cha Zhang, Wanxiang Che, et~al.
\newblock Layoutlmv2: Multi-modal pre-training for visually-rich document
  understanding.
\newblock In {\em Proceedings of the 59th Annual Meeting of the Association for
  Computational Linguistics and the 11th International Joint Conference on
  Natural Language Processing (Volume 1: Long Papers)}, pages 2579--2591, 2021.

\bibitem{yang2022code}
Ziyi Yang, Yuwei Fang, Chenguang Zhu, Reid Pryzant, Dongdong Chen, Yu Shi,
  Yichong Xu, Yao Qian, Mei Gao, Yi-Ling Chen, et~al.
\newblock i-code: An integrative and composable multimodal learning framework.
\newblock {\em arXiv preprint arXiv:2205.01818}, 2022.

\bibitem{zhong2019publaynet}
Xu Zhong, Jianbin Tang, and Antonio~Jimeno Yepes.
\newblock Publaynet: largest dataset ever for document layout analysis.
\newblock In {\em 2019 International Conference on Document Analysis and
  Recognition (ICDAR)}, pages 1015--1022. IEEE, 2019.

\end{thebibliography}
}

\appendix

\begin{figure*}[t]
  \centering
  \includegraphics[width=0.995\textwidth]{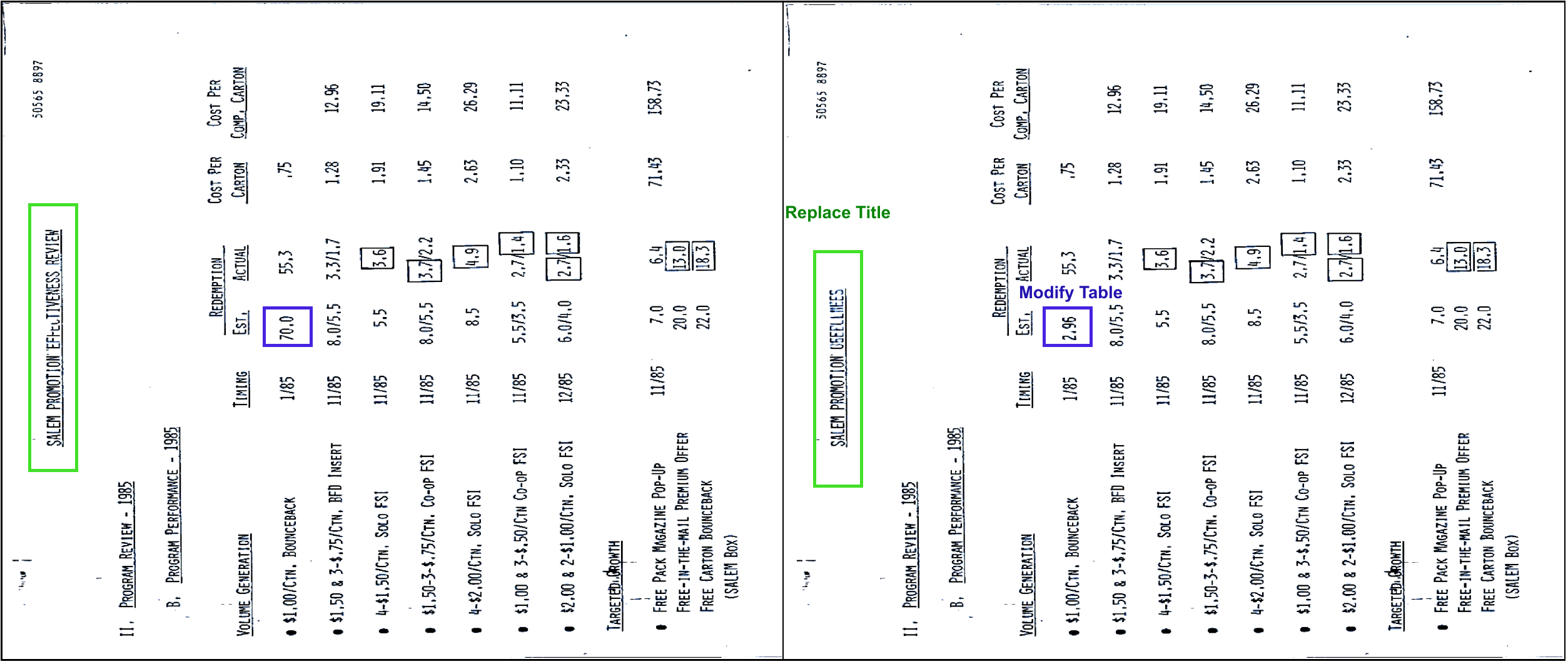}
  \includegraphics[width=0.995\textwidth]{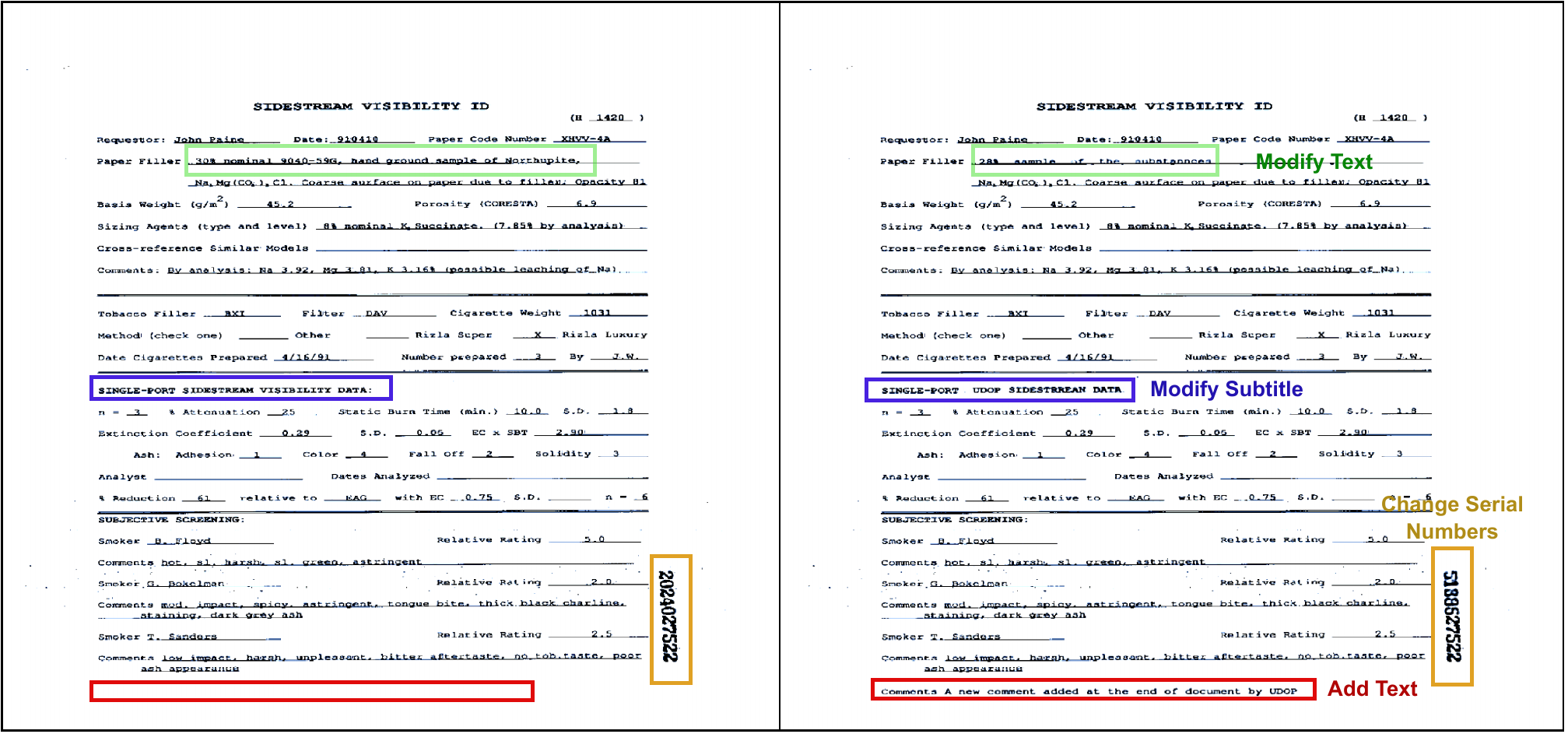}
  \includegraphics[width=0.995\textwidth]{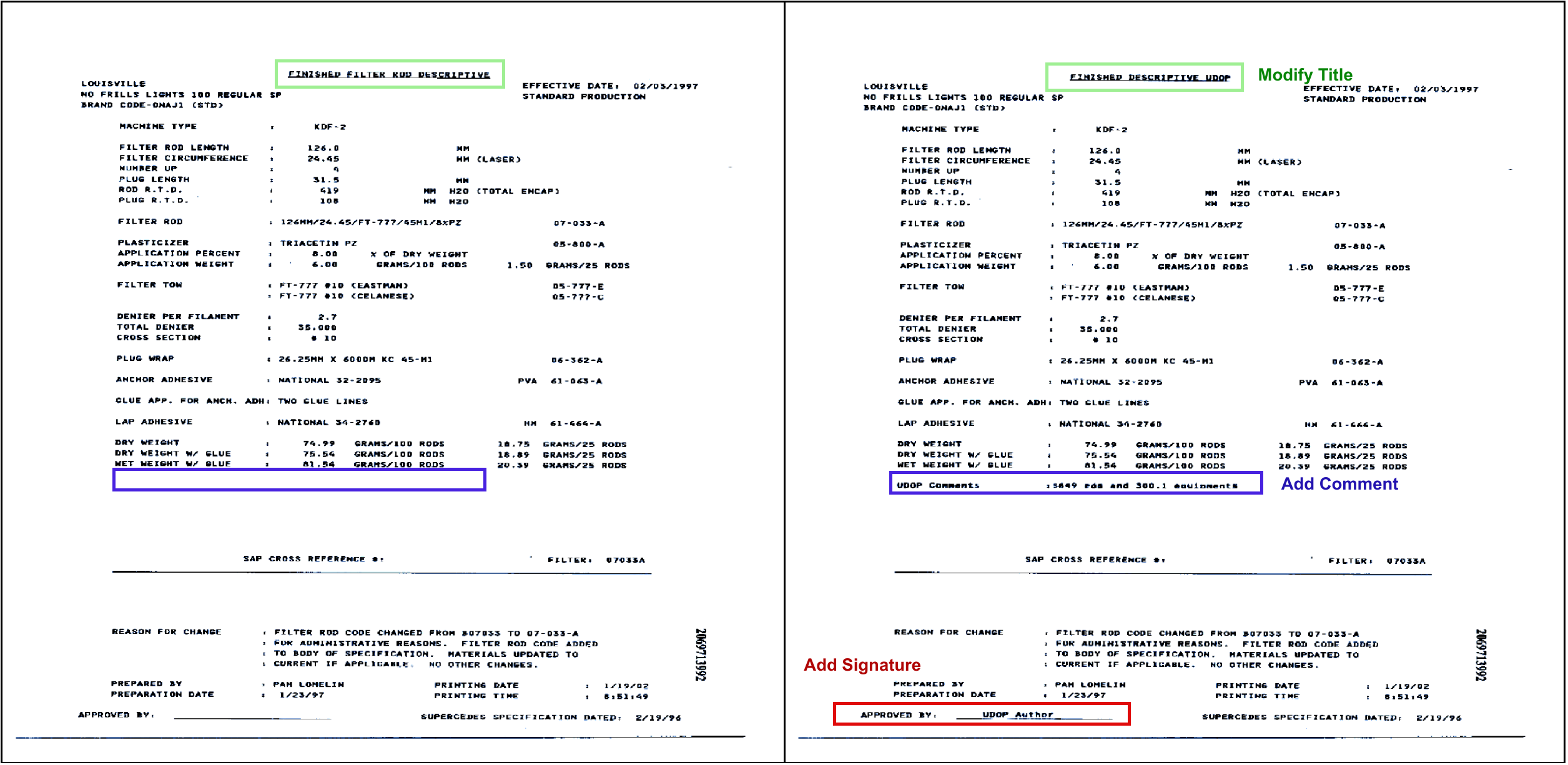}
  \caption{
  Document generation with customized content (right). Left is the original document. We show different document edits within the same figure including title replacement, text addition, text replacement, and tilted text replacement. All edits are done with one model run.
}
\label{fig:vis_mae_creative2}
\end{figure*}

\begin{figure*}[t]
  \centering
  \includegraphics[width=0.8\textwidth]{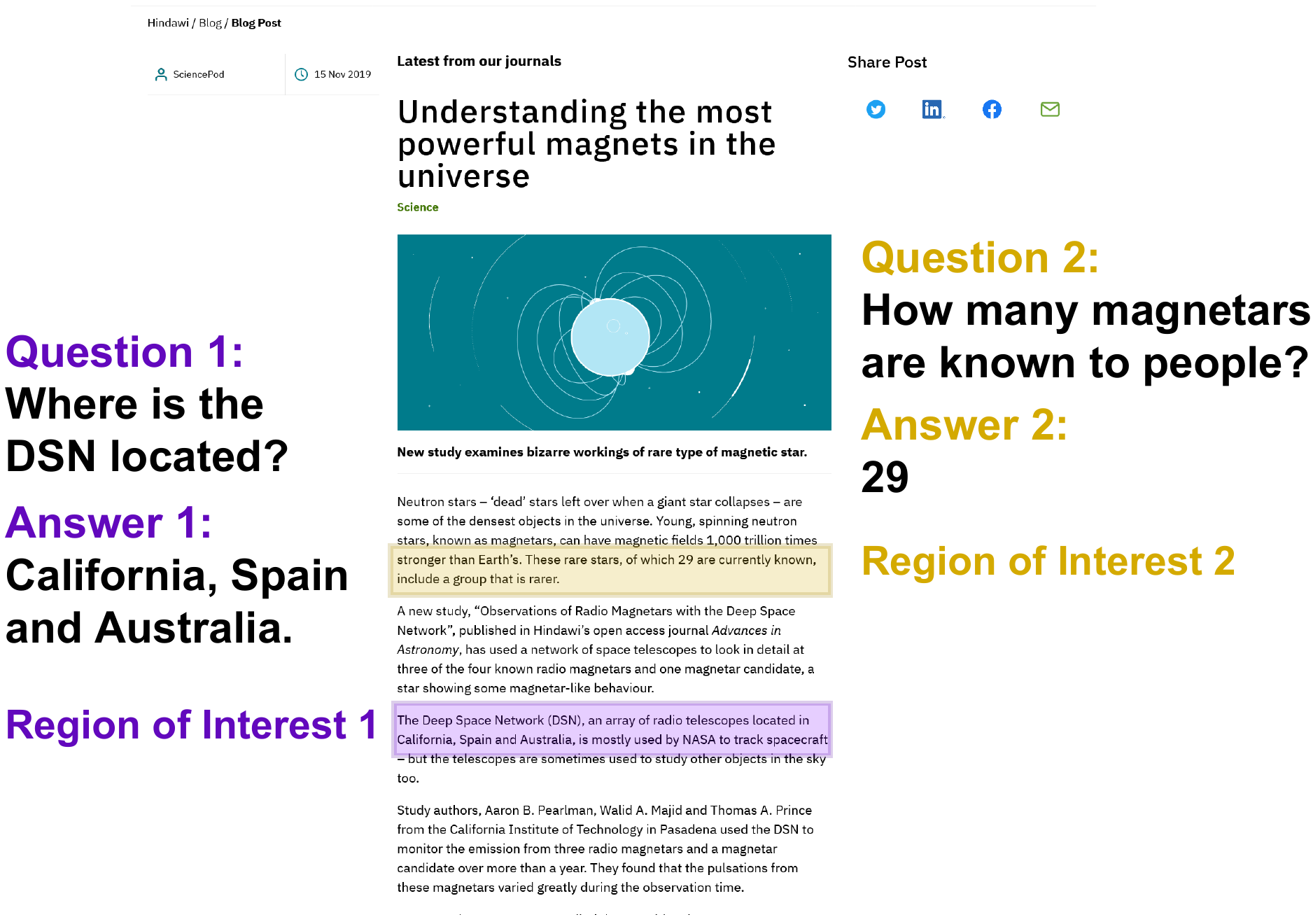}
  \caption{
  Document QA and answer localization with \methodname{} on  VisualMRC dataset. As shown, besides generating the answer, \methodname{} can predict the region of interest (RoI) that answer is located in by generating the layout tokens. Note that the the labeled RoI VisualMRC dataset is at paragraph level. 
}
\label{fig:vis_qa}
\end{figure*}

\begin{table*}[h]
\caption{
Comparison of different image size in curriculum learning on the \duebenchmark{}. Modality T, L, V denote text, layout, or vision.}
\label{tab:cl_due_results}
\centering
\resizebox{0.8\textwidth}{!}{
\begin{tabular}{l c cc ccc cc c}
\toprule
\multirow{2}{*}{Model} & \multirow{2}{*}{Modality} & \multicolumn{2}{c}{Question Answering} &  \multicolumn{3}{c}{Information Extraction} &  \multicolumn{2}{c}{Table QA/NLI} & \multirow{2}{*}{Avg.} \\
\cmidrule(lr){3-4} \cmidrule(lr){5-7} \cmidrule(lr){8-9} 
& & \docvqa{} & InfoVQA & KLC & \pwc{} & \deepform{} & \wtq{} & \tabfact{} &\\
\midrule
\textbf{\methodname{} (224)} & V+T+L & 84.4 & 46.1 & 82.1 & 26.7 & 83.6 & 46.1 & 78.2 & 63.9 \\
\textbf{\methodname{} (512)} & V+T+L & 84.5 & 47.3 & 82.0 & 27.1 & 84.7 & 46.2 & 78.3 & 64.3 \\
\textbf{\methodname{} (1024)} & V+T+L & \textbf{84.7} & \textbf{47.4} & \textbf{82.8} & \textbf{28.9} & \textbf{85.5} & \textbf{47.2} & 78.9 & \textbf{65.1} \\
\bottomrule
\end{tabular}
}
\end{table*}

\begin{table*}[t]
\caption{
Performance with standard deviations on on the \duebenchmark{}. Modality T, L, V denote text, layout, or vision.}
\label{tab:var_due_results}
\centering
\resizebox{\textwidth}{!}{
\begin{tabular}{l c cc ccc cc c}
\toprule
\multirow{2}{*}{Model} & \multirow{2}{*}{Modality} & \multicolumn{2}{c}{Question Answering} &  \multicolumn{3}{c}{Information Extraction} &  \multicolumn{2}{c}{Table QA/NLI} & \multirow{2}{*}{Avg.} \\
\cmidrule(lr){3-4} \cmidrule(lr){5-7} \cmidrule(lr){8-9} 
& & \docvqa{} & InfoVQA & KLC & \pwc{} & \deepform{} & \wtq{} & \tabfact{} &\\
\midrule
Donut & V & 72.1 & - & - & - & - & - & - & - \\
BERT\textsubscript{large}~\cite{devlin2018bert} & T & 67.5 & - & - & - & - & - & - & - \\
T5\textsubscript{large}~\cite{raffel2020exploring} & T & 70.4 & 36.7 & 74.3 & 25.3 & 74.4 & 33.3 & 58.9 & 50.7 \\
T5\textsubscript{large}+U~\cite{powalski2021going} & T & 76.3 & 37.1 & 76.0 & 27.6 & 82.9 & 38.1 & 76.0 & 56.5 \\
T5\textsubscript{large}+2D~\cite{powalski2021going} & T+L & 69.8 & 39.2 & 72.6 & 25.7 & 74.0 & 30.8 & 58.0 & 50.4 \\
T5\textsubscript{large}+2D+U~\cite{powalski2021going} & T+L & 81.0 & 46.1 & 75.9 & 26.8 & 83.3 & 43.3 & 78.6 & 59.8 \\
LAMBERT~\cite{garncarek2021lambert} & T+L & - & - & 81.3 & - & - & - & - & - \\
StructuralLM\textsubscript{large}\cite{li2021structurallm} & T+L & 83.9 & - & - & - & - & - & - & - \\
LayoutLMv2\textsubscript{large}~\cite{xu2021layoutlmv2} & V+T+L & 78.8 & - & - & - & - & - & - & - \\
LayoutLMv3\textsubscript{large}~\cite{huang2022layoutlmv3} & V+T+L & 83.4 & 45.1 & 77.1 & 26.9 & 84.0 & 45.7 & 78.1 & 62.9 \\ 
\textbf{\dual{}} & V+T+L & 84.4$\pm$0.1 & 47.1$\pm$0.2 & 81.9$\pm$0.4 & 28.7$\pm$0.5 & 85.2$\pm$0.2 & 46.7$\pm$0.4 & \textbf{79.5}$\pm$0.3 & 64.7$\pm$0.3 \\
\textbf{\methodname{}} & V+T+L & \textbf{84.7}$\pm$0.2 & \textbf{47.4}$\pm$0.2 & \textbf{82.8}$\pm$0.3 & \textbf{28.9}$\pm$0.4 & \textbf{85.5}$\pm$0.2 & \textbf{47.2}$\pm$0.2 & 78.9$\pm$0.1 & \textbf{65.1}$\pm$0.2 \\
\bottomrule
\end{tabular}
}
\end{table*}
\begin{table*}[t]
\caption{
Performance with standard deviations on \funsd{}, \cord{}, and \rvlcdip{} datasets.
}
\label{tab:var_main_results}
\centering
\begin{tabular}{l c cccc}
\toprule
\multirow{2}{*}{Model} & \multirow{2}{*}{Modality} & \multicolumn{2}{c}{Info Ext.} &  \multicolumn{1}{c}{Classification} \\
\cmidrule(lr){3-4} \cmidrule(lr){5-5}
 & & \funsd{} & \cord{} & \rvlcdip{} \\
\midrule
Donut & V & - & 91.6 & 95.3 \\
BERT\textsubscript{large} & T & 65.63 & 90.25 & 89.92 \\
BROS\textsubscript{large}\cite{hong2022bros} & T+L & 84.52 & 97.40 & - \\
StructuralLM\textsubscript{large} & T+L & 85.14 & - & 96.08 \\
LiLT\cite{wang2022lilt} & T+L & 88.41 & 96.07 & 95.68 \\
FormNet~\cite{lee2022formnet} & T+L & 84.69 & 97.28 & - \\
LayoutLM\textsubscript{large} & T+L & 77.89 & - & 91.90 \\
SelfDoc & V+T+L & 83.36 & - & 92.81 \\
UDoc & V+T+L & 87.93 & 98.94 & 95.05 \\
DocFormer\textsubscript{large}~\cite{appalaraju2021docformer} & V+T+L & 84.55 & 96.99 & 95.50 \\
TILT\textsubscript{large} & V+T+L & - & 96.33 & 95.52 \\
LayoutLMv2\textsubscript{large} & V+T+L & 84.20 & 96.01 & 95.64 \\
LayoutLMv3\textsubscript{large} & V+T+L & 92.08 & 97.46 & 95.93\\
\textbf{\dual{}} & V+T+L & 91.20$\pm$0.21 & 97.64$\pm$0.12 & 96.22$\pm$0.27 \\
\textbf{\methodname{}} & V+T+L & 91.62$\pm$0.34 & 97.58$\pm$0.15 & 96.00$\pm$0.26 \\
\bottomrule
\end{tabular}
\end{table*}

\section{Appendix Overview}
The appendix has the following contents:
\begin{itemize}
\item Vision demonstrations of \methodname{} localizing answers in documents, the effectiveness of the cross attention with character embeddings in vision generation, and more neural editing examples \Cref{supp:demos}.
\item \dual{} performance in \Cref{sec:dual_perf}.
\item More details for pretraining and evaluation datasets, and finetuning experiment set up in \Cref{supp:ss_tasks}.
\item Experiment results of curriculum learning in \Cref{supp:cl}.
\item Performance variance of \methodname{} in \Cref{supp:variance}.
\item Discussion of limitations and societal impacts in \Cref{supp:limitations}.

\end{itemize}

\section{Visualization Analysis}
\label{supp:demos}

\noindent \textbf{Creative Image Generation.} \methodname{} achieves controllable high-quality document generation and editing as described in \Cref{sec:vis_analysis}. We show additional examples here in \cref{fig:vis_mae_creative2}. Our model can edit and add to the document image content with customized contents. Note that even if the document content is vertical (the first subfigure of \cref{fig:vis_mae_creative2}), \methodname{} can still achieve high generation quality.

\noindent \textbf{Answer Localization for Document QA.} \methodname{} can perform question answering while predicting the location of the answer. We show examples on \visualmrc{} in \Cref{fig:vis_qa} and our model can answer the questions regarding the document correctly while locating the area of interest.

\section{\dual{} Performance}
We list the performance of \dual{} on \funsd{}, \cord{}, and \rvlcdip{} in \Cref{tab:udop_apped}.
\label{sec:dual_perf}
\begin{table}[t]
\caption{
Performance of \dual{} on \funsd{}, \cord{}, and \rvlcdip{}.
}
\label{tab:udop_apped}
\centering
\resizebox{1.01\columnwidth}{!}{
\begin{tabular}{l c cccc}
\toprule
\multirow{2}{*}{Model} & \multirow{2}{*}{Modality} & \multicolumn{2}{c}{Info Ext.} &  \multicolumn{1}{c}{Classification} \\
\cmidrule(lr){3-4} \cmidrule(lr){5-5}
 & & \funsd{} & \cord{} & \rvlcdip{} \\
\midrule
Donut~\cite{kim2021donut} & V & - & 91.6 & 95.3 \\
BERT\textsubscript{large}~\cite{devlin2018bert} & T & 65.63 & 90.25 & 89.92 \\
BROS\textsubscript{large}~\cite{hong2022bros} & T+L & 84.52 & 97.40 & - \\
StructuralLM\textsubscript{large}~\cite{li2021structurallm} & T+L & 85.14 & - & 96.08 \\
LiLT\cite{wang2022lilt} & T+L & 88.41 & 96.07 & 95.68 \\
FormNet~\cite{lee2022formnet} & T+L & 84.69 & 97.28 & - \\
LayoutLM\textsubscript{large}~\cite{xu2020layoutlm} & T+L & 77.89 & - & 91.90 \\
SelfDoc~\cite{li2021selfdoc} & V+T+L & 83.36 & - & 92.81 \\
UniDoc~\cite{gu2021unidoc} & V+T+L & 87.93 & 96.86 & 95.05 \\
DocFormer\textsubscript{large}~\cite{appalaraju2021docformer} & V+T+L & 84.55 & 96.99 & 95.50 \\
TILT\textsubscript{large}~\cite{powalski2021going} & V+T+L & - & 96.33 & 95.52 \\
LayoutLMv2\textsubscript{large}~\cite{xu2021layoutlmv2} & V+T+L & 84.20 & 96.01 & 95.64 \\
LayoutLMv3\textsubscript{large}~\cite{huang2022layoutlmv3} & V+T+L & \textbf{92.08} & 97.46 & 95.93\\
\textbf{\dual{}} & V+T+L & 91.20 & 97.64 & \textbf{96.22} \\
\textbf{\methodname{}} & V+T+L & 91.62 & \textbf{97.58} & 96.00 \\
\bottomrule
\end{tabular}
}
\end{table}

\section{Supervised Pretraining Tasks}
\label{supp:ss_tasks}
In this section, we list more details about the supervised datasets in pretraining and evaluations.

\subsection{Classification}
\rvlcdip{}~\cite{harley2015evaluation} contains 16 document categories, such as ``invoice'', ``scientific publication'' and ``form''. The dataset has 320k training, 40k validation and 40k test images. 

\subsection{Layout Analysis}
\publaynet{}~\cite{zhong2019publaynet} is a layout analysis dataset created from medical publications. It contains over 360k document images and labeled with typical document layout elements such as titles, paragraphs, etc.

\subsection{Information Extraction}
\docbank{}~\cite{li2020docbank} is a richly-annotated large-scale IE dataset. It consists of 500K document pages, where 400K for training, 50K for validation and 50K for testing. It has 12 semantic structure labels like abstract, title, and author. Each token has corresponding bounding box and semantic structure label.

\kleister{}~\cite{stanislawek2021kleister} is an IE dataset with complex invoice page layout and has 21.6k entities and 2.7k document images from UK Charity
Commission. Its entities for extraction include invoice date, 
invoice number, net amount, vendor name, etc.

\pwc{}~\cite{kardas2020axcell} is an IE dataset which has 2,291 leaderboards, where the data is collected from the Papers with Code labelling interface. It asks information like task, dataset, metric, etc. Different from original implementation, \duebenchmark{} provides complete papers as input instead of tables.

\deepform{}~\cite{svetlichnaya2020deepform} is an IE dataset collected from political television ads in US elections and has 20k receipts and over 100k document images. This task is to extract entities like advertiser name, contract number, amount paid, etc. 
\subsection{Question Answering}
\websrc{}~\cite{chen2021websrc} stands for Web-based Structural Reading Comprehension. It consists of 0.44M questions collected from 6.5K web pages with corresponding HTML, screenshots and metadata. The answer is either the text span of context or yes/no.

\visualmrc{}~\cite{tanaka2021visualmrc} stands for visual machine reading comprehension. It consists of 10,197 images 30,562 abstractive questions-answers.

\docvqa{}~\cite{mathew2021docvqa} is a QA dataset for excerpts from industry documents and has 50k questions on 12k document images. It asks questions on topics like text content, non-textual elements like marks or diagrams, layout, style, etc.

\infovqa{}~\cite{mathew2022infographicvqa} is a QA dataset with a focus on infographic images and has 30K questions on 5.3k document images. It requires reasoning on text content, images, data visualizations, layout, etc.

\wtq{}~\cite{pasupat2015compositional} is a table-based QA dataset on HTML tables collected from Wikipedia. It has 2.1k tables and 22k questions hand crafted by humans and cover a wide range of topics like table lookup, superlatives, arithmetic operations, etc.

\subsection{Document NLI}

\tabfact{}~\cite{chen2019tabfact} is an open-domain table-based NLI task and has 16k Wikipedia tables for 118k statements by human annotations.

\subsection{Finetuning Experiment Setting}
\label{sec:apd_finetune}
For all \duebenchmark{} finetuning experiments, we use Adam~\cite{kingma2014adam} optimizer with learning rate 5e-5, 1000 warmup steps, batch size 16, weight decay of 1e-2, $\beta_1=0.9$, and $\beta_2=0.98$. For \funsd{} and \cord{}, we use learning rate 3e-4 and for \rvlcdip{}, we use learning rate 1e-3 both with 1000 warmup steps, batch size 16, weight decay of 1e-2, $\beta_1=0.9$, and $\beta_2=0.98$.

\section{Curriculum Learning}
\label{supp:cl}
In this section, we present the results of curriculum learning of input image resolution (224, 512, 1024) on the validations sets of evaluation benchmarks. As shown in \Cref{tab:cl_due_results}, while the model already performs competitively well on 224 resolution, its performance further increases on 512 and 1024.

\section{Performance Variance}
\label{supp:variance}
For results in \Cref{tab:due_results} and \Cref{tab:main_results}, we report their standard deviations as shown in \Cref{tab:var_due_results} and \Cref{tab:var_main_results}. The deviations are computed from 5 runs with different seeds for parameter initialization.

\section{Limitations and Societal Impact}
\label{supp:limitations}

\methodname{} can assist users with document analysis, understanding and information extraction. This automatic processing technology will make the document processing workflow more efficient and potential more accurate. It is also worth noting that, similar to all AI generation technology, the document generation capacity of \methodname{} can be potentially abused for malicious document counterfeit, e.g., signature forgery, tampering monetary amount in checks, fake medical/financial records generation, etc. To avoid abuse, for model release we plan to open source the vision generation model only with limited access, e.g., through an API. Documents submitted by users that are classified as sensitive (the classifier can be a finetuned \methodname{} model), such as checks and personal ID, will be denied.

Applying \methodname{} on non-English data, especially those with non-Latin writing systems, may require further modifications to the model. For example, in \cref{sec:ss_tasks}, the vision decoder cross-attends with character embeddings. Then for non-English data, we need to include more character embeddings to attend with.

\end{document}